\documentclass[10pt,twocolumn,letterpaper]{article}

\usepackage[pagenumbers]{iccv} 
\usepackage{xspace}
\usepackage[T1]{fontenc}
\usepackage{stfloats}
\usepackage{booktabs}
\usepackage{graphicx}
\usepackage{tcolorbox}
\usepackage{listings}
\tcbuselibrary{breakable}
\definecolor{iccvblue}{rgb}{0.21,0.49,0.74}
\usepackage[pagebackref,breaklinks,colorlinks=true,citecolor=iccvblue]{hyperref}
\usepackage{caption}
\usepackage{multirow}
\usepackage{float}

\newcommand{\method}{LMM-R1\xspace}

\def\paperID{14177} 
\def\confName{ICCV}
\def\confYear{2025}

\title{\includegraphics[height=1.5em]{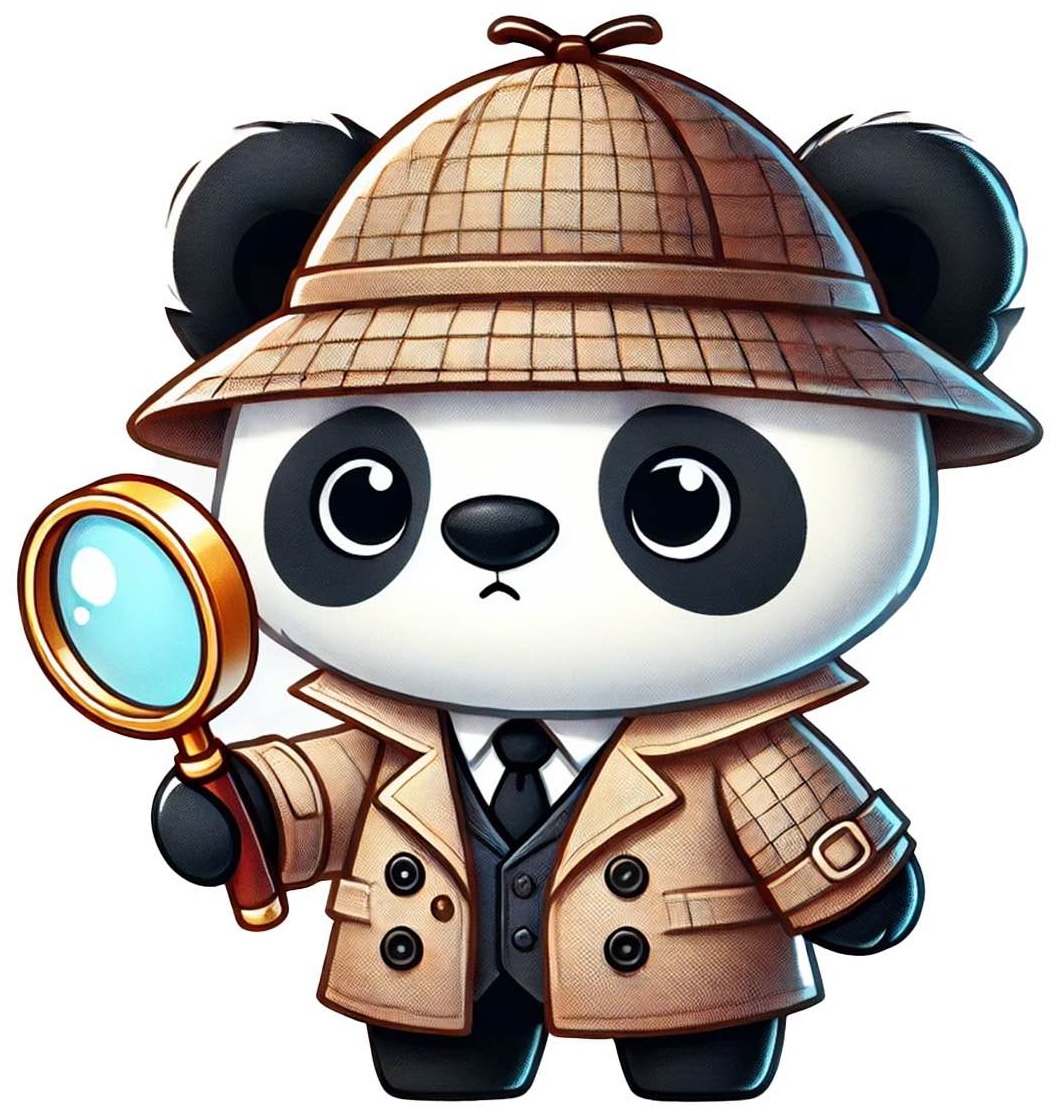} LMM-R1: Empowering 3B LMMs with Strong Reasoning Abilities Through Two-Stage Rule-Based RL}

\author{Yingzhe Peng$^{1,4,*}$ \and
Gongrui Zhang$^{1,*}$ \and
Miaosen Zhang$^{1,*}$ \and
Zhiyuan You$^{2}$ \and
Jie Liu$^{2}$ \and
Qipeng Zhu$^{3}$ \quad
Kai Yang$^{4}$ \quad
Xingzhong Xu$^{4}$ \quad 
Xin Geng$^{1}$ \quad Xu Yang$^{1,\ddagger}$\\[2ex]
$^{1}$Key Laboratory of New Generation Artificial Intelligence Technology and \\
Its Interdisciplinary Applications (Southeast University), Ministry of Education, China\\
$^{2}$The Chinese University of Hong Kong \quad $^{3}$Fudan University \quad $^{4}$Ant Group\\
{\tt\small \{yingzhe.peng, zgr, miazhang, yangxu\_palm\}@seu.edu.cn}\\
{\tt\small \url{https://github.com/TideDra/lmm-r1}}
}

\begin{document}

\maketitle

\footnotetext[1]{$^{*}$Equal contribution. $^{\ddagger}$Corresponding author. Yingzhe Peng works done during his internship at Ant Group and his mentor is Kai Yang.}

\begin{abstract}
    Enhancing reasoning in Large Multimodal Models (LMMs) faces unique challenges from the complex interplay between visual perception and logical reasoning, particularly in compact 3B-parameter architectures where architectural constraints limit reasoning capacity and modality alignment. 
    While rule-based reinforcement learning (RL) excels in text-only domains, its multimodal extension confronts two critical barriers: (1) data limitations due to ambiguous answers and scarce complex reasoning examples, and (2) degraded foundational reasoning induced by multimodal pretraining. 
    To address these challenges, we propose \textbf{\method}, a two-stage framework adapting rule-based RL for multimodal reasoning through \textbf{Foundational Reasoning Enhancement (FRE)} followed by \textbf{Multimodal Generalization Training (MGT)}. The FRE stage first strengthens reasoning abilities using text-only data with rule-based RL, then the MGT stage generalizes these reasoning capabilities to multimodal domains. 
    Experiments on Qwen2.5-VL-Instruct-3B demonstrate that \method achieves 4.83\% and 4.5\% average improvements over baselines in multimodal and text-only benchmarks, respectively, with a 3.63\% gain in complex Football Game tasks. These results validate that text-based reasoning enhancement enables effective multimodal generalization, offering a data-efficient paradigm that bypasses costly high-quality multimodal training data. 
\end{abstract}    
\vspace{-12pt}
\section{Introduction}
\label{sec:intro}
\vspace{-3pt}

\begin{figure}[!t]
\centering
\includegraphics[width=1.0\columnwidth]{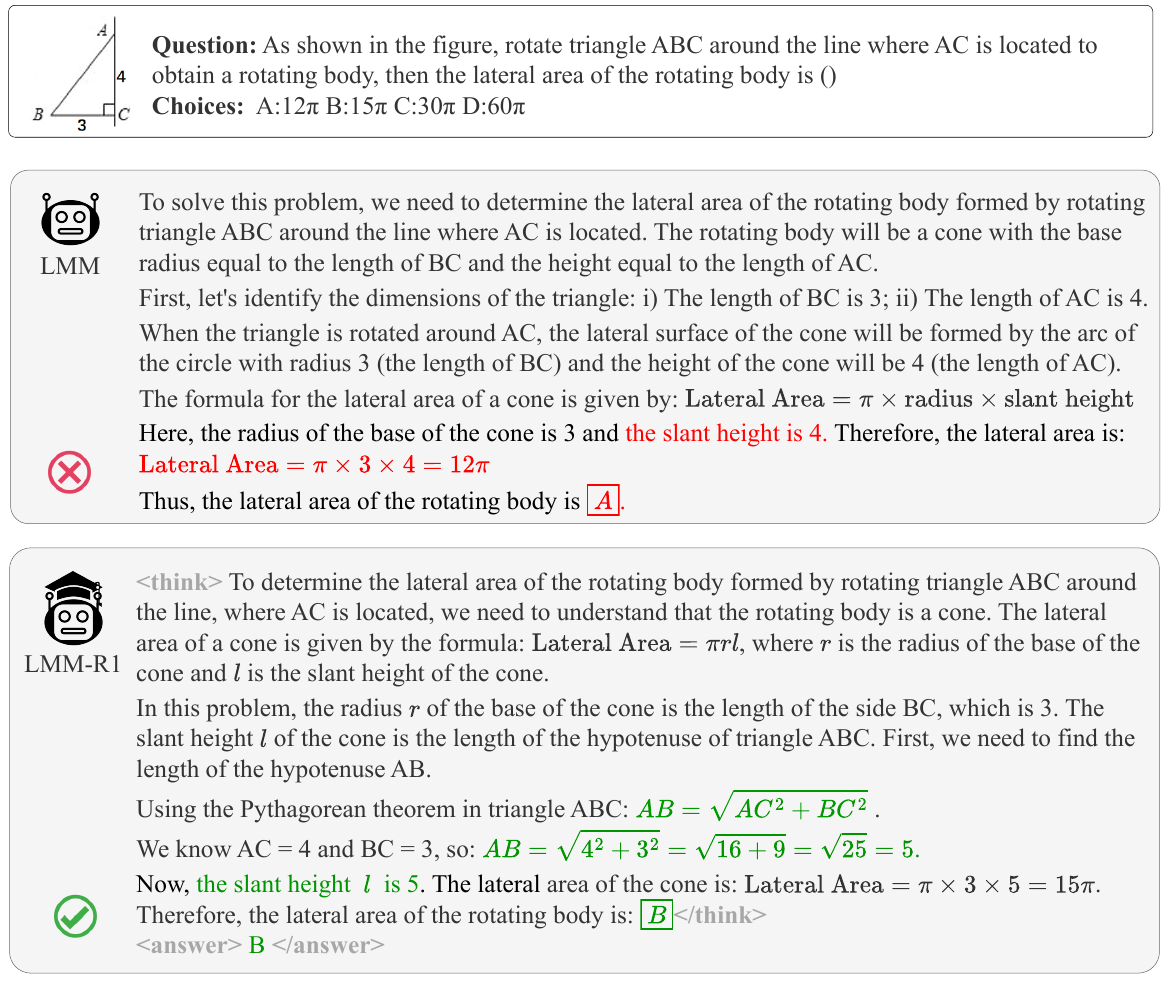}
\vspace{-17pt}
\caption{Comparison of reasoning approaches on a geometric problem. The baseline LMM (top) incorrectly identifies the slant height, leading to an erroneous solution. In contrast, LMM-R1 (bottom) demonstrates superior reasoning by correctly applying the Pythagorean theorem to determine the actual slant height, showcasing how our rule-based RL approach enhances the model's ability to apply proper mathematical principles.}
\label{fig:teaser}
\vspace{-20pt}
\end{figure}

Recent years have witnessed remarkable progress in Large Language Reasoning Models (LLRMs)~\cite{deepseekr1,o1,qwq}. Researchers have explored approaches like Monte Carlo Tree Search (MCTS)~\cite{mcts_dpo, omega_prm, mulberry} and Process Reward Models (PRM)~\cite{stepbystep, pav, omega_prm}, which require extensive human annotation~\cite{stepbystep} or expensive computational resources~\cite{mcts_dpo, omega_prm, mathshepherd}. DeepSeek-R1~\cite{deepseekr1} introduced a more efficient approach with Rule-based Reinforcement Learning (RL), which only requires prompt-answer pairs, enabling models to autonomously improve reasoning through exploration while avoiding reward hacking~\cite{deepseekr1}. When we turn our attention to the multimodal domain, the challenges of enhancing reasoning capabilities become considerably more complex. 
Large Multimodal Models (LMMs) face greater challenges as visual information increases reasoning complexity by requiring integration of both visual perception and logical reasoning. This challenge is particularly severe for 3B LMMs, as their limited parameter capacity constrains their capabilities. 
Moreover, such requirement makes high-quality multimodal reasoning data extremely difficult to collect~\cite{lmm_reason_survey}, further complicating the use of multimodal data with reasoning processes for training~\cite{mathvision, mathverse, olympiadbench}.

Given rule-based RL's potential in the text-only domain~\cite{simplerl-reasoning,deepscaler}, we aim to extend it to multimodal reasoning. However, direct application faces two specific issues: (1) \textbf{Data limitations}: Rule-based RL requires uniquely verifiable answers for accurate rewards, yet multimodal tasks often involve answer ambiguity (\eg, image description, visual QA), while also suffering from an abundance of perception-focused data but limited complex reasoning examples, which may lead to insufficient reasoning in RL; (2) \textbf{Weak foundational reasoning}: Models trained on multimodal data often show weakened capabilities on text-only tasks~\cite{vila}, and some LMMs using Chain-of-Thought (CoT) actually experience performance degradation on multimodal benchmarks~\cite{comt, atomthink}, a phenomenon amplified in smaller 3B-parameter architectures due to their limited capacity. 

\begin{figure*}[htb]
\centering
\includegraphics[width=\textwidth]{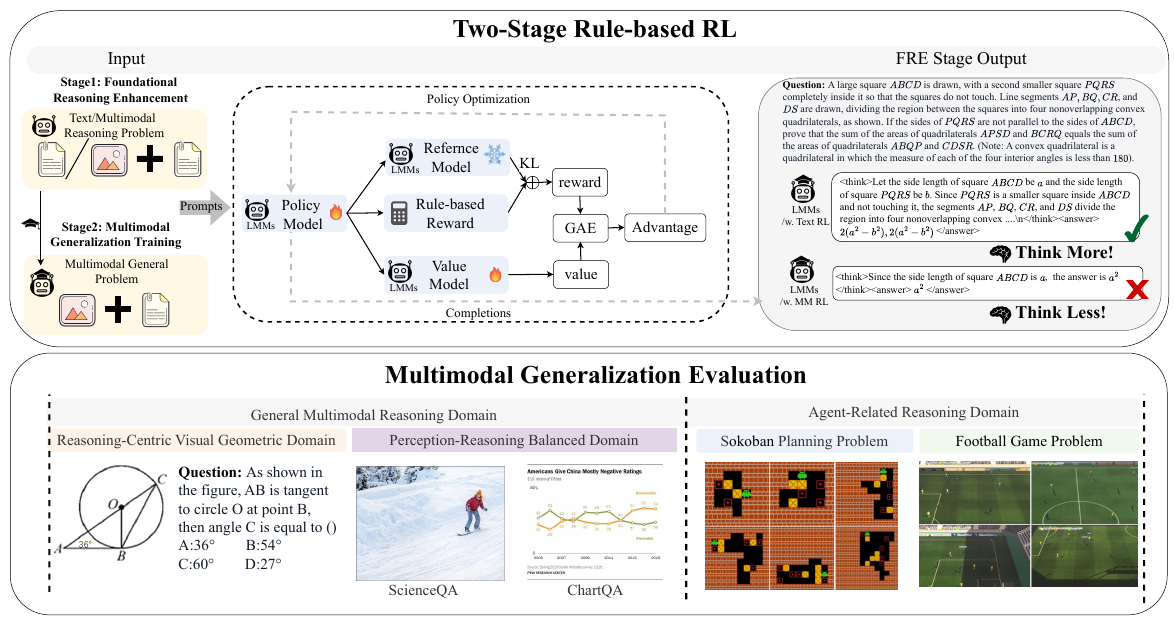}
\vspace{-17pt}
\caption{The top shows that \method uses a two-stage, rule-based RL strategy to enhance reasoning capabilities. The first Foundational Reasoning Enhancement (FRE) stage trains LMM using text-only reasoning data to improve its foundational reasoning skills. Subsequently, the second Multimodal Generalization Training (MGT) stage then extends these capabilities across diverse multimodal domains, with evaluation benchmarks illustrated in the bottom panel.}
\label{fig:pipeline}
\vspace{-15pt}
\end{figure*}

To address these challenges, we propose a simple yet effective two-stage rule-based RL training framework: \textbf{\method}, aimed at enhancing the general reasoning capabilities of LMMs. While previous studies suggest that the smaller 3B LMM inherently lacks the capacity for complex multimodal reasoning, we demonstrate that strategic two-stage training can overcome these architectural constraints. As shown in \cref{fig:teaser}, LMM-R1 correctly applies the Pythagorean theorem to determine the slant height of the cone, while the baseline LMM incorrectly identifies it, demonstrating how our approach enhances mathematical reasoning capabilities. Our framework consists of two key stages. The first stage is \textbf{Foundational Reasoning Enhancement (FRE)}, which uses rule-based RL with abundant high-quality \textit{text-only} data to strengthen the model's basic reasoning abilities.
This stage establishes a solid reasoning foundation that serves as a crucial stepping stone for subsequent multimodal generalization, avoiding the need for expensive multimodal data. 
The second stage is \textbf{Multimodal Generalization Training (MGT)}, where we continue rule-based RL training on limited complex multimodal reasoning tasks. This continuing training generalizes the reasoning abilities to various multimodal domains. In the MGT stage, we focus on two key multimodal reasoning domains: general multimodal reasoning domain and agent-related reasoning domain. For general multimodal reasoning domain, we further extend the model's reasoning abilities to various multimodal scenarios including GeoQA, ScienceQA, and ChartVQA, \etc.

For the agent-related domain, which represent significant real-world use cases of LMMs~\cite{visualagentbench, mobileagentv2}, we evaluate our approach on tasks such as Sokoban~\cite{sokoban} and football tasks~\cite{football}.
These tasks require sophisticated reasoning skills, including goal identification, path planning, and multi-image processing, thus provide meaningful assessments of real-world applications. Besides, continuing training the model on Sokoban with rule-based RL further improves its performance on the agent-related benchmark, further validating \method's effectiveness in agent domains.

In experiments, we use Qwen2.5-VL-Instruct-3B~\cite{qwen2.5vl} as the baseline model and apply LMM-R1 on it and the results reveal several important findings. First, enhancing foundational reasoning capabilities is crucial for multimodal reasoning. Using text-only data for RL training can significantly improve the multimodal reasoning capabilities both in general domain and agent-related domain, while directly using multimodal data for rule-based RL yields limited improvement in reasoning abilities. As shown in \cref{fig:pipeline}, our experiments reveal that the model fails to generate high-quality long reasoning processes after directly trained on verifiable multimodal data with rule-base RL. 
Furthermore, \method not only enhances the model's reasoning capabilities but also further improves its visual abilities. Specifically, \method achieves significant performance improvements across multiple multimodal benchmarks, such as a 4.5\%/4.83\%/3.63\% performance increase over the baseline model on the text-only/five multimodal/ Football Game benchmarks. Our main contributions include:
\begin{itemize}
    \item We introduce \method, the first framework using rule-based RL with a two-stage training strategy (FRE and MGT) to enhance multimodal reasoning without extensive human annotation.
    \item We show that using text-only reasoning data through rule-based RL can largely improve the foundational reasoning ability and more importantly, such ability can be generalized to multimodal domains.
    \item We demonstrate that even for a 3B LMMs which initially possess very limited reasoning capabilities, \method can significant enhance the reasoning ability, suggesting the vast potential of \method for broader applications.
\end{itemize}

\section{Related Work}
\label{sec:related_work}

\noindent\textbf{Large Multimodal Model (LMM).}
LMMs integrate additional modalities, particularly vision, into Large Language Models (LLMs)~\cite{gpt4, llama, vicuna, qwen} to enhance general vision capabilities. Initially, models like Flamingo~\cite{flamingo} and BLIP-2~\cite{blip2} aligned frozen vision encoders with LLMs for visual question answering. Subsequently, the LLaVA series~\cite{llava, llava1.5, llavanext}, MiniGPT-4~\cite{minigpt4}, and mPLUG-Owl series~\cite{mplugowl, mplugowl2, mplugowl3} introduced visual instruction tuning to improve instruction-following. Models such as VisionLLM~\cite{visionllm}, KOSMOS-2~\cite{kosmos2}, Shikra~\cite{shikra}, and the Qwen-VL series~\cite{qwenvl, qwen2vl, qwen2.5vl} enhanced LMMs with visual grounding for tasks like region description and localization. InternVL~\cite{internvl} scaled up vision foundation models for alignment with LLMs. Additionally, GPT-4V~\cite{gpt4v} and Gemini~\cite{gemini} demonstrated strong general visual understanding. Mixture-of-Experts (MoE) approaches improved understanding in DeepSeek-VL2~\cite{deepseekvl2}, Uni-MoE~\cite{unimoe}, MoVA~\cite{mova}, and MoME~\cite{mome}. Some models, including SEED-X~\cite{seedx}, Chameleon~\cite{chameleon}, Show-o~\cite{showo}, Transfusion~\cite{transfusion}, and Janus~\cite{janus, januspro}, unified vision understanding and generation. However, most existing LMMs still lack reasoning capabilities.

\noindent\textbf{Reinforcement Learning in LLMs and LMMs.}
Reinforcement Learning (RL) has become a key methodology for enhancing the capabilities of LLMs and LMMs. Early research primarily focused on Reinforcement Learning from Human Feedback (RLHF)~\cite{ouyang2022training}, which aimed to align model outputs with human preferences~\cite{ouyang2022training,bai2022training,sun2023aligning}. Recent advancements have demonstrated that RL can significantly enhance the reasoning capabilities of these models. For instance, models such as DeepSeek-R1~\cite{deepseekr1} and Kimi-1.5~\cite{team2025kimi} highlight the effectiveness of RL in improving LLMs' reasoning abilities through rule-based reward. In the multimodal domain, research on leveraging RL to enhance LMMs' reasoning capacities remains in its early stages. Some researchers~\cite{ursa, atomthink} explore using Process Reward Models (PRM) to enhance the LMMs' reasoning capabilities. However, these PRM-based approaches typically require powerful closed-source models to generate large amounts of training data, resulting in high computational and financial costs. Concurrent work R1-V~\cite{chen2025r1v} explores rule-based RL in specific subdomains such as geometry problems and object counting tasks, but lacks exploration in general domains and agent-related applications.
\section{Preliminaries}

\subsection{Reinforcement Learning for LMMs}

We use the Proximal Policy Optimization (PPO) algorithm~\cite{ppo} to train the LMMs, which aims to maximize the following objective function:
\begin{equation}
    \hspace{-5pt} 
    \label{eq:ppo_objective}
    \setlength{\abovedisplayskip}{3pt}
    \setlength{\belowdisplayskip}{3pt}
    \mathcal{L}(\theta) = \mathbb{E}_{y \sim \pi_{\theta}}[r(y) - \beta \cdot \text{KL}(\pi_{\theta}(y | \mathcal{I}, x) || \pi_{\theta_0}(y | \mathcal{I}, x))],
\end{equation}
where $y$ is the generated answer, $\mathcal{I}$ is the image input (if any), $x$ is the text prompt, $\pi_{\theta}$ is the policy model, $\pi_{\theta_0}$ is the fixed initial policy, and $\beta$ is the Kullback-Leibler divergence (KL) penalty coefficient. 

\subsection{Reward Function}
We follow ~\cite{deepseekr1} in designing a two-part reward function for rule-based RL:

\noindent\textbf{Format Reward.} We first evaluate whether the responses follow the required structured format, \ie, wrapping the reasoning within \texttt{<think></think>} tags, followed by the final answer in \texttt{<answer></answer>} tags. This format encourages the model to explicitly demonstrate its reasoning process before providing the final answer, which is crucial for readability and evaluating the reasoning quality.

\noindent\textbf{Accuracy Reward.} The second metric evaluates the correctness of the solution. We employ a symbolic verification approach~\cite{mathshepherd} that parses both the model's answer and the ground-truth solution into comparable representations. These representations are then checked for equivalence rather than exact string matching, allowing for different but equivalent expressions to be recognized as correct.

The final reward function \( r(y) \) can be described simply as follows:
\(
r(y) = \alpha \cdot r_f(y) + r_a(y)
\).
Here, \( r_f(y) \) is the format reward, and \( r_a(y) \) is the accuracy reward. The parameter \( \alpha \) adjusts how important the format is compared with the accuracy.
This reward function can effectively guide the model not only to produce correct solutions, but also to articulate its reasoning process in a structured manner.

\section{\method: Two-Stage Rule-based RL}
\label{sec:methods}
Inspired by the approach used to develop DeepSeek-R1 from DeepSeek-R1-Zero~\cite{deepseekr1}, we divide our multimodal reasoning model training into two stages: (1) Increase the model's foundational reasoning ability with rule-based RL using high-quality text-only; (2) Generalize reasoning ability across three distinct and complex multimodal reasoning tasks independently.

\subsection{Foundational Reasoning Enhancement (FRE)}
To enhance the foundational reasoning ability of the base model, we explore two complementary approaches:

\noindent\textbf{Text-Only Reasoning Enhancement:} We utilize large-scale and high-quality verifiable text-only data for rule-based RL. This approach leverages a wide variety of text-based reasoning problems, which are inherently more challenging and demand more complex reasoning processes compared to existing multimodal reasoning tasks. By training on these rich textual reasoning tasks, we aim to develop strong foundational reasoning capabilities for the model, which can be potentially transferred to multimodal contexts.

\noindent\textbf{Multimodal Reasoning Enhancement:} As a comparison, we also utilize available multimodal verifiable data for rule-based RL training. While this data is more limited in the quality, this approach provides the model with direct exposure to the multimodal domain. As a result, it offers more immediate benefits for multimodal reasoning tasks and enhances the model's ability to understand visual contexts.

\subsection{Multimodal Generalization Training (MGT)}
After enhancing the foundational reasoning capabilities of the model, we focus on evaluating how well these capabilities generalize to diverse multimodal domains. We evaluate across two distinct domains that test different aspects of multimodal intelligence: General Multimodal Reasoning Domain and Agent-Related Reasoning Domain. By continuing RL training on these domains, we aim to improve the multimodal reasoning ability of the model.

\subsubsection{General Multimodal Reasoning Domain}

This domain focuses on the visual perception ability to perform reasoning based on both image contents and texts.

\noindent\textbf{Visual Reasoning-Centric Geometric Domain (Geo):} We select geometric reasoning as our first domain for continued RL training due to its natural bridge between our initial reasoning enhancement stage and multimodal applications. This domain shares structural similarities with the mathematical reasoning used in FRE stage while introducing visual perception challenges. By continuing RL training on geometric problems, the model learns to extract visual information and apply mathematical reasoning. This domain is relatively simple, as the model already possesses rich pre-trained knowledge about geometric concepts, and the data distribution is similar to that of the FRE stage.

\noindent\textbf{Perception-Reasoning Balanced Domain (PerceReason):} For our second continued RL training domain, we employ a broad spectrum of multimodal tasks from over 20 distinct datasets that require more perception ability than geometric domain, including visual question answering, document understanding, mathematical reasoning, and scientific reasoning. This diverse training domain exposes the model to various visual contexts and reasoning problems simultaneously. By continuing RL training across this comprehensive collection, the model must adapt its reasoning capabilities to handle heterogeneous inputs and tasks—mirroring real-world application scenarios. This domain presents a moderate challenge since the model has rich pretrained multimodal knowledge, but the data distribution differs significantly from the FRE stage. This stage evaluates whether enhancing reasoning ability by FRE stage training builds a strong foundation model. The foundation model is expected to be efficiently transferred to various challenging multimodal domain via continued reinforcement learning.

\subsubsection{Agent-Related Reasoning Domain}

To evaluate and enhance the model's ability to act as an agent in complex visual environments, we follow MageBench~\cite{magebench} to evaluate the how well our enhanced reasoning capabilities transfer to tasks requiring sequential decision-making and planning in visual contexts. We select two domains from MageBench in the following.

\noindent\textbf{Sokoban Planning Domain:} Sokoban is a classic puzzle game that requires the agent to push boxes to designated target locations. This domain evaluates the model's spatial reasoning and planning capabilities, as it must visualize potential moves, anticipate deadlocks, and generate an optimal sequence of actions. 

\noindent\textbf{Football Game Domain:} This domain places the model in a multi-agent football environment where it must control a player to achieve game objectives. The model needs to make strategic decisions based on the positions of teammates and opponents, demonstrating cooperation and interaction skills in a competitive setting.

In addition, we also continually train the model on Sokoban via rule-based RL to enhance the model's agent capabilities, further validating our approach's effectiveness in agent domains. Notably, these domains represent scenarios that the model has not encountered during pretraining and serve as a challenging test of generalization, as they have data distributions significantly different from our first-stage rule-based RL training.

\section{Experiments}
\label{sec:experiments}

\begin{table*}[th]
    \centering
    \setlength{\tabcolsep}{5pt}
    \caption{
    Results (\%) across benchmarks categorized by three reasoning intensities: High-Level Reasoning (Text-Only) (MATH500/GPQA), Multimodal Reasoning (OlympiadBench/MathVision/MathVerse), and General Multimodal (MM-Star/MathVista). The "MM Avg" column displays the average performance across all multimodal benchmarks. The \textbf{best} result is \textbf{bolded} and the \underline{second-best} is \underline{underlined}.
    }
    \label{tab:benchmark_results}
    \vspace{-5pt}
    \begin{tabular}{l|ccc|ccc|cc|c}
        \toprule
        \multirow{2}{*}{\textbf{Model}} & \multicolumn{3}{c|}{\textbf{Text-Only}} & \multicolumn{3}{c|}{\textbf{MM Reasoning-Dominated}} & \multicolumn{2}{c|}{\textbf{MM General}} & \multirow{2}{*}{\textbf{MM Avg}} \\
        \cmidrule{2-9}
        & \textbf{MATH} & \textbf{GPQA} & \textbf{Avg} & \textbf{Olymp.} & \textbf{MathVis.} & \textbf{MathVer.} & \textbf{MM-Star} & \textbf{MathVista} & \\
        \midrule
        Qwen2.5-VL CoT & 63.40 & 30.30 & 46.85 & 10.28 & 23.59 & 34.64 & 51.40 & 60.70 & 36.12 \\
        \midrule
        \multicolumn{10}{c}{\textit{Foundational Reasoning Enhancement Stage}} \\
        \midrule
        FRE-Multi & 61.80 & 27.27 & 44.54 & 11.80 & 24.74 & 38.45 & \textbf{58.76} & \textbf{64.20} & 38.71 \\
        FRE-Text & \underline{65.40} & \underline{36.87} & \underline{51.14} & \textbf{15.62} & 25.76 & 38.83 & 55.15 & 61.40 & \underline{39.35} \\
        \midrule
        \multicolumn{10}{c}{\textit{Multimodal Generalization Training Stage}} \\
        \midrule
        MGT-Geo & \textbf{65.80} & 32.32 & 49.06 & \underline{14.63} & \textbf{26.84} & \textbf{41.80} & 54.39 & 59.00 & 39.33 \\
        MGT-PerceReason & 63.80 & \textbf{38.89} & \textbf{51.35} & \textbf{15.62} & \underline{26.35} & \underline{41.55} & \underline{58.03} & \underline{63.20} & \textbf{40.95} \\
        \bottomrule
    \end{tabular}
    \vspace{-10pt}
\end{table*}

In the experiments, we use Qwen2.5-VL-Instruct-3B~\cite{qwen2.5vl} as our baseline model and apply \method framework on it. First, by training the baseline model with rule-based RL in multimodal and text-only datasets respectively, we get the reasoning-enhanced foundation model: FRE-Multi and FRE-Text (FRE stage). Then we continue the rule-based RL training on the FRE-* model in other multimodal domains (MGT stage) and get the MGT-Geo, MGT-PerceReason model and MGT-Sokoban. Besides, we also train baseline model with rule-based RL by using the same datasets of different domains and name these models as Direct-RL-* (e.g., Direct-RL-Geo, Direct-RL-Sokoban) for comparison with our MGT-* models.

\subsection{Setting and implementation details}
\noindent\textbf{Datasets:} We use multiple datasets across different training stages in our experiments.

\noindent\textbf{FRE Stage:} We use the following datasets for the FRE stage to enhance the reasoning capability of the model: \textbf{(1) Text-Only RL Dataset:} We use the prompt-answer pairs from DeepScaler-Preview-Dataset~\cite{deepscaler}, which contains 40k math prompt-answer pairs and all of them have unique verifiable answers. \textbf{(2) Verifiable Multimodal-65K (VerMulti):} We use the MathV360K~\cite{mathv} dataset, which contains 360k multimodal problems with diverse domains, including arithmetic, geometry, calculus, science, and more. The dataset is further filtered, saving prompts that have verifiable numeric answers (\eg, $123$, $4.11$) or option answer (\eg, A, B, C, D). Finally, we get 130k multimodal problems. To reduce the cost of training, we randomly sample 65k data from the filtered dataset for rule-based RL training. 

\noindent\textbf{MGT Stage:} We continue training FRE-Text model with rule-based RL using different datasets for different domains. \textbf{(1). Visual Reasoning-Centric Geometry Domain:} We extract the geometry problems from VerMulti, collecting 15K geometry problems and getting the VerMulti-Geo.  \textbf{(2). Perception-Reasoning Balanced Domain:} We directly use the VerMulti dataset. \textbf{(3). Sokoban Domain:} We follow ~\cite{sokoban} to randomly generate 11500 environments for sokoban game as our training set.

See implementation details in \cref{app:training_details}.

\subsection{Experiments Results}
We evaluate the performance of the model from two main aspects: General Task Benchmarks in \cref{sec:basic_general_results} and Agent-Related Benchmarks in \cref{sec:agent_general_results}. General Task Benchmarks focus on the model's fundamental capabilities in reasoning, visual perception, and multimodal understanding, which serve as the foundation for handling complex tasks. In contrast, Agent-Related Benchmarks concentrate on assessing the model's performance in planning, decision-making, and goal-oriented tasks, which are essential abilities for building intelligent systems capable of autonomous operation in real-world environments.

\subsubsection{General Task Benchmarks}
\label{sec:basic_general_results}

To comprehensively evaluate our models' capabilities, we conduct experiments across both multimodal and text reasoning benchmarks. For multimodal evaluation, we use five benchmarks as categorized in \cref{tab:benchmark_results}: OlympiadBench~\cite{olympiadbench}, MathVision~\cite{mathvision} test, and MathVerse~\cite{mathverse} testmini for reasoning-dominated tasks, while MM-Star~\cite{mmstar} and MathVista~\cite{mathvista} testmini assess general multimodal capabilities. Complementing this, we evaluate pure reasoning abilities using MATH500~\cite{stepbystep} and GPQA~\cite{gpqa} datasets. MATH500 consists of 500 mathematics problems covering algebra, geometry, probability, and calculus, whereas GPQA comprises 448 multiple-choice questions in biology, physics, and chemistry authored by domain experts. This dual evaluation approach comprehensively assesses the reasoning ability of the model in various scenario ranging from basic text-only context and challenging multimodal context

\noindent\textbf{Foundational Reasoning Enhancement Analysis}
Our evaluation across both multimodal and text-only reasoning benchmarks reveals distinct performance patterns between FRE-Multi and FRE-Text models, highlighting important trade-offs in multimodal model training.

On \textbf{Text-Only Benchmarks}, the divergence becomes even more pronounced. FRE-Text shows substantial improvements over the baseline, with a 2.0\% increase on MATH500 and 6.57\% on GPQA, resulting in a 4.29\% overall enhancement in text-only performance. In contrast, FRE-Multi exhibits performance declines of 1.6\% on MATH500 and 3.03\% on GPQA, suggesting that training the model with relatively simple multimodal data may compromise its complex reasoning abilities, despite the improvement in visual reasoning ability.

On \textbf{Multimodal Benchmarks}, both models demonstrate improvements over the baseline Qwen2.5-VL, but with different strengths. For general multimodal tasks, FRE-Multi shows substantial gains with a 3.5\% improvement on MathVista and 7.36\% on MM-Star, while FRE-Text exhibits more modest improvements of 0.7\% and 3.75\%, respectively. This suggests that RL with multimodal data more effectively enhances general visual capabilities. Conversely, for multimodal reasoning-dominated tasks, FRE-Text demonstrates stronger performance with improvements of 5.34\% on OlympiadBench, 2.17\% on MathVision, and 4.19\% on MathVerse, compared to FRE-Multi's smaller gains of 1.52\%, 1.15\%, and 3.81\%, respectively.

These results reveal a fundamental trade-off: multimodal RL enhances vision-related capabilities but may compromise pure reasoning abilities, whereas text-only RL strengthens core reasoning capabilities that effectively transfer to multimodal reasoning contexts while providing more limited benefits for tasks primarily requiring visual perception. This performance dichotomy underscores the importance of training data selection in developing models with balanced capabilities.

Based on these findings, we select FRE-Text as our reasoning-enhanced foundation for subsequent experiments, as it provides the strongest foundational reasoning capabilities while maintaining competitive performance on multimodal tasks, offering the best platform for further domain-specific enhancement.

\begin{table*}[ht]
    \centering
    \caption{
    Results (\%) on geometry-related benchmarks. For MathVision, results are reported for Analytic/Combinatorial/Metric/Solid Geometry. For MathVerse, results are categorized by modality emphasis: TD (Text Domain)/TL (Text Lite)/VI (Vision Intensive)/VD (Vision Domain)/VO (Vision Only). The \textbf{best} performance in each subfield is \textbf{bolded}.
    }
    \label{tab:continue_geo_rl_results}
    \vspace{-5pt}
    \begin{tabular}{l|ccccc|cccccc}
        \toprule
        \multirow{2}{*}{\textbf{Model}} & \multicolumn{5}{c|}{\textbf{MathVision}} & \multicolumn{6}{c}{\textbf{MathVerse}} \\
        \cmidrule{2-12}
        & \textbf{Analy.} & \textbf{Combin.} & \textbf{Metric} & \textbf{Solid} & \textbf{AVG} & \textbf{TD.} & \textbf{TL.} & \textbf{VI.} & \textbf{VD.} & \textbf{VO.} & \textbf{Avg} \\
        \midrule    
        Qwen2.5-VL CoT & 34.52 & 20.78 & 26.33 & 20.49 & 25.53 & 43.15 & 35.41 & 33.38 & 32.87 & 28.43 & 34.64\\
        Direct-RL-Geo & 30.95 & 17.53 & 26.59 & 22.54 & 24.40 & 47.59 & 40.36 & 38.96 & 36.04 & 27.03 & 38.00\\
        FRE-Text & 28.57 & 22.08 & 31.01 & 24.10 & 26.44 & 48.22 & 42.26 & \textbf{39.72} & 38.96 & 25.00 & 38.83\\
        MGT-Geo & \textbf{36.90} & \textbf{22.73} & \textbf{31.66} & \textbf{27.87} & \textbf{29.79} & \textbf{51.02} & \textbf{42.51} & \textbf{39.72} & \textbf{39.09} & \textbf{36.68} & \textbf{41.80} \\
        \bottomrule
    \end{tabular}
    \vspace{-10pt}
\end{table*}

\vspace{2pt}\noindent\textbf{Multimodal Generalization Training Analysis}

\vspace{2pt}\noindent\textbf{Geo Domain:} We continue rule-based RL training on the VerMulti-Geo dataset to obtain the MGT-Geo model. As shown in \cref{tab:benchmark_results}, MGT-Geo surpasses the baseline by 3.21\% on multimodal benchmarks, demonstrating strong multimodal capabilities.

To evaluate geometry-specific performance, we analyze results on geometry subsets from MathVision (categorized by mathematical geometry subfields: Analytic, Combinatorial, Metric, and Solid, which are the in-domain questions for VerMulti-Geo) and MathVerse (classified by visual dependency: from Text Domain to Vision Only). For comparison, we also train the baseline, Qwen2.5-VL, with the same VerMulti-Geo dataset by rule-based RL training method and get the Direct-RL-Geo, which skips the FRE stage. Results are presented in \cref{tab:continue_geo_rl_results}.

From \cref{tab:continue_geo_rl_results}, Direct-RL-Geo improves performance on simpler MathVerse tasks by 3.36\% but decreases performance on more complex MathVision by 1.13\% compared to the baseline, indicating limitations in developing robust reasoning through domain-specific training alone. While MGT-Geo achieves superior performance across all benchmarks, with significant improvements over FRE-Text on MathVision by 3.35\% and MathVerse by 2.97\%. 

A particularly revealing finding emerges when examining the ``Vision Only'' category in MathVerse. Here, FRE-Text shows a 3.43\% decline compared to the baseline, indicating that text-only reasoning enhancement may come at the cost of visual perception capabilities. However, MGT-Geo not only recovers but significantly enhances these capabilities, achieving an 8.25\% improvement over the baseline and an impressive 11.68\% improvement over FRE-Text. Interestingly, Direct-RL-Geo actually decreases performance in this category by 1.4\%, further highlighting the importance of the FRE stage.

\noindent\textbf{PerceReason Domain: }Inspired by the success of the MGT-Geo model, we continue the rule-based RL training on the Perception-Reasoning Balanced Domain with VerMulti. The comprehensive benchmark results are shown in \cref{tab:benchmark_results}. The MGT-PerceReason model shows substantial performance improvements compared to FRE-Text, with an average increase of 1.6\% across multimodal benchmarks. 

This improvement is particularly pronounced in benchmarks that emphasize general visual tasks: the model achieves a 1.8\% gain on MathVista and a 2.88\% improvement on MM-Star. Notably, these enhancements in visual capabilities do not come at the expense of reasoning performance. In fact, MGT-PerceReason maintains or even improves upon the strong reasoning capabilities of FRE-Text across text-only and reasoning-focused benchmarks.

These results provide compelling evidence that our proposed two-stage training framework creates a powerful synergistic effect. The FRE stage provides a stronger platform for domain-specific training, enabling simultaneous improvement in both reasoning and visual perception capabilities that cannot be achieved through either approach alone. By first enhancing fundamental reasoning capabilities through complex text-only data before introducing multimodal data, this approach effectively augments the model's performance on general visual tasks while preserving its core reasoning abilities. Importantly, this two-stage training strategy successfully circumvents the performance degradation in reasoning tasks that typically occurs when models are trained directly with multimodal RL data (as observed with the FRE-Multi model). The consistent improvements across diverse benchmarks demonstrate that \textbf{\textit{FRE stage provides a more robust platform for developing multimodal models that excel in both perception-heavy and reasoning-intensive tasks, effectively boosting multimodal intelligence through an efficient training paradigm.}}

\subsubsection{Agent Domain Generalization}
\label{sec:agent_general_results}

One of the most intriguing applications for LMM is agents, which involves more complex reasoning. Specifically, it requires reasoning and perceiving interleaved process. 
We select Sokoban as a test scenario for rule-based RL training, as it is easy to scale and verify. 
The model is required to observe only the initial state and then, after deliberation, provide the complete sequence of actions in a single output.
We randomly initialize 10,000 simple difficulty Sokoban levels and trained using the PPO algorithm. For comparison, MageBench~\cite{magebench} provides several environments and level settings for Sokoban that enable fair comparison. During training, as before, we use the rule-based task reward from MageBench, along with a minor format reward. The training results are shown in \cref{tab:agent_result}.

During testing, we evaluate not only the results of Sokoban-Global, which shares the same environment and agent settings as the training phase, but also test an unseen environment with completely different settings in MageBench, specifically the Football-Online results. The last block in \cref{tab:agent_result} presents the original QwenVL-2.5 results, as well as the results from text-only training discussed earlier. The results indicate that the reasoning capabilities derived from text-only training exhibit stable generalization in agent environments.
Training within the corresponding in-domain environment can demonstrate further performance improvements. Using a model with only 3B parameters, after training, it can match or even surpass product-level large models.
This suggests that rule-based reinforcement learning may have significant potential applications in the fields of agent and embodied AI planning.

\begin{table}[t]
  \centering
  \setlength\tabcolsep{4pt}
  \caption{
  Agent-level evaluation on MageBench-Sokoban and Football environments. All values represent well-defined rewards. \textbf{Global setting}: The model observes the initial scene once and determines all subsequent actions. \textbf{Online setting}: The model observes the game scene and makes one action per time step.
  }
    \vspace{-7pt}
    \begin{tabular}{lrr}
    \toprule
    Model & \multicolumn{1}{c}{Sokoban-Global} & \multicolumn{1}{c}{Football-Online} \\
    \midrule
    Idle baseline & \multicolumn{1}{c}{41.18} & \multicolumn{1}{c}{2.53} \\
    Human level & \multicolumn{1}{c}{83.63} & \multicolumn{1}{c}{54.68} \\
    \midrule
    Claude-3.5-Sonnet & \multicolumn{1}{c}{48.26} & \multicolumn{1}{c}{16.94} \\
    Gemini-1.5-pro & \multicolumn{1}{c}{46.13} & \multicolumn{1}{c}{18.33} \\
    GPT-4o & \multicolumn{1}{c}{46.09} & \multicolumn{1}{c}{21.20} \\
    \midrule
    Qwen2.5-VL-3B & \multicolumn{1}{c}{42.35}      &  \multicolumn{1}{c}{15.36} \\
    FRE-Text &   \multicolumn{1}{c}{44.81 (+2.46)}    & \multicolumn{1}{c}{18.46} \\
    Direct-RL-Sokoban &   \multicolumn{1}{c}{46.89 (+4.54)}     & \multicolumn{1}{c}{16.32}  \\
    MGT-Sokoban &       \multicolumn{1}{c}{\textbf{47.91 (+5.56)}}  & \multicolumn{1}{c}{\textbf{18.99}} \\
    \bottomrule
    \end{tabular}%
    \label{tab:agent_result}
  \vspace{-15pt}
\end{table}%

\subsection{Discussion}
\noindent\textbf{Generalization Capability of RL Compared to SFT}

\vspace{2pt}\noindent To investigate the generalization capabilities of different training paradigms, we conduct a comparative analysis between RL and SFT approaches during the first training stage. For this experiment, we use identical text datasets, DeepScaler-40K~\cite{deepscaler}, for both training methodologies. For the SFT approach, we first generate comprehensive reasoning processes for each query-answer pair using DeepSeek-R1-Distill-Qwen-14B~\cite{deepseekr1}, then employ these structured triplets, \ie, (query, reasoning process, answer), as training data to obtain SFT-Text. Subsequently, we fine-tune the SFT-Text model on the VerMulti-Geo dataset using rule-based RL to produce SFT-Geo. For the RL approach, we use FRE-Text and MGT-Geo as the comparative models to SFT-Text and SFT-Geo, respectively.

\begin{figure}[t]
     \centering
     \subfloat[General Reasoning Benchmark]{
     \includegraphics[height=0.75\columnwidth]{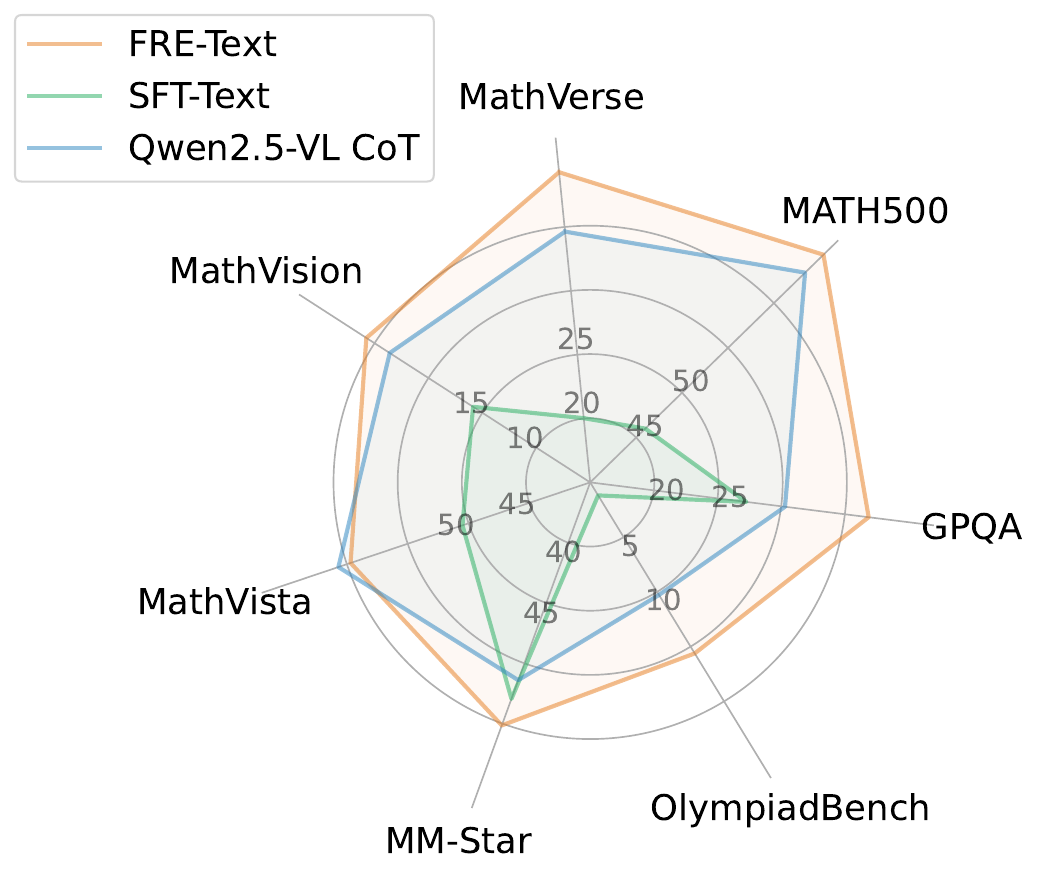}
     }
     \vfill
     \subfloat[Geometry Specialized Benchmark]{
     \includegraphics[height=0.75\columnwidth]{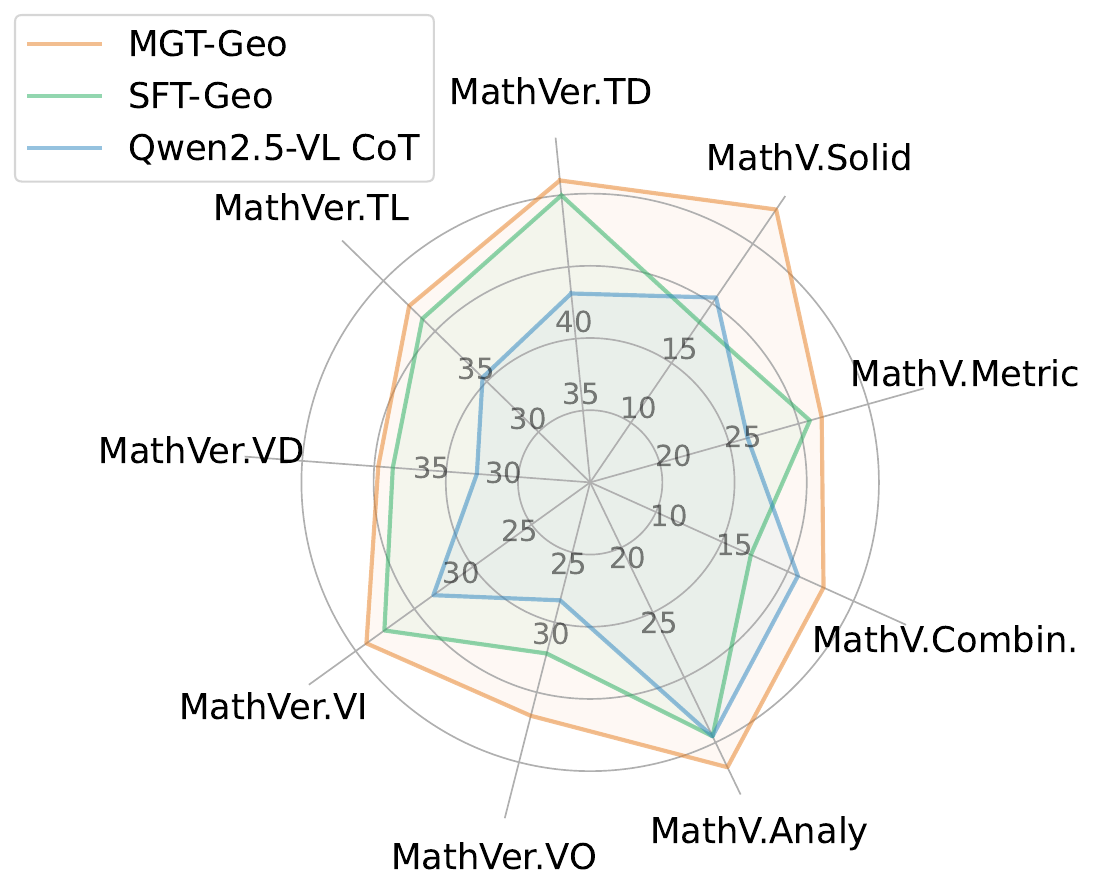}
     }
     \vspace{-5pt}
     \caption{Results (\%) on geometry domain with SFT and RL.}
     \label{fig:geometry_domain_results}
     \vspace{-5pt}
\end{figure}

We evaluate both training paradigms on general reasoning capabilities and geometry domain generalization performance, with results shown in \cref{fig:geometry_domain_results}. Our findings reveal that direct SFT with text data on the baseline model leads to significant performance degradation, exhibiting catastrophic forgetting~\cite{forgetting}. In contrast, using the same data for RL training results in substantial performance improvements. Furthermore, when evaluated on downstream tasks, SFT-Geo begins to recover performance on geometry tasks, with overall downstream performance exceeding the baseline but still falling short of the RL model. These results demonstrate that compared to SFT, RL not only avoids catastrophic forgetting but also more effectively transfers reasoning capabilities to other domains.

\vspace{2pt}\noindent\textbf{Why Multimodal RL is not better than Text-only RL?}

\vspace{2pt}\noindent Our experimental results reveal a counterintuitive phenomenon: while multimodal RL training enhances performance on multimodal benchmarks, it simultaneously leads to a decline in text-only reasoning capabilities compared to the baseline model. To elucidate this paradox, we conduct a detailed analysis of model behavior during the RL training process across different datasets, examining both output patterns and reasoning strategies.

\begin{figure}[t]
    \centering
    \includegraphics[width=0.8\columnwidth]{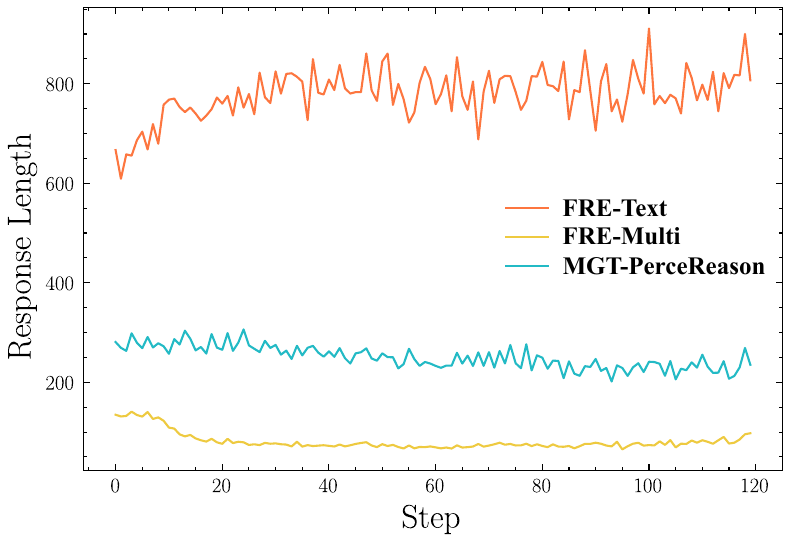}
    \vspace{-10pt}
    \caption{Response length trends of 3 models during RL training.}
    \label{fig:cast_study_in_rl}
    \vspace{-15pt}
\end{figure}

We show the response length trending of models trained on different datasets exhibit striking divergence in \cref{fig:cast_study_in_rl}. FRE-Multi shows a consistent downward trend in response length, decreasing from approximately 150 tokens at initialization to below 80 tokens after 120 training steps. In contrast, FRE-Text demonstrates rapid growth in response length during the initial training phase, rising from 600 tokens to over 800 tokens within the first 40 steps, after which it maintains a stable length around 800 tokens for the remainder of the training period. While MGT-PerceReason demonstrates a particularly interesting pattern in \cref{fig:cast_study_in_rl}. After FRE stage training, its response length stabilizes between 200-250 tokens during subsequent multimodal training, maintaining a balance between reasoning depth and visual recognition efficiency.

Furthermore, we analyze sampled outputs across different training stages to elucidate this divergence. Comparing model outputs between the initial model and after 120 training steps reveals distinct patterns. The FRE-Multi model exhibits a clear trend toward brevity: while the initial model produces relatively detailed reasoning, by 120 steps it generates significantly condensed responses that directly identify visual elements with minimal reasoning steps, despite maintaining answer accuracy. Conversely, the FRE-Text model demonstrates the opposite trajectory: its responses after 120 steps become more elaborate compared to the initial model, incorporating additional reasoning steps, explicit mathematical formulations, and more comprehensive explanations. This pattern aligns with the quantitative length measurements presented in \cref{fig:cast_study_in_rl}. Detailed examples of these contrasting behaviors are given in \cref{app:case_study}.

The results suggest that the two-stage training strategy is a practical and advantageous approach for developing robust multimodal reasoning models, especially given the current scarcity of high-quality multimodal reasoning data.

\section{Conclusion}
\label{sec:conclusion}
We introduce \method, a framework designed to enhance multimodal reasoning in LMMs through rule-based RL. This is achieved via a two-stage strategy: Foundational Reasoning Enhancement (FRE) and Multimodal Generalization Training (MGT). Despite the initial limitations in reasoning capabilities of 3B LMMs, \method substantially improves their performance. Experiments show that text-only reasoning establishes a solid foundation for multimodal generalization, creating a synergistic effect that outperforms models trained solely on text or multimodal data. In agent-related domains, \method notably enhances performance in complex planning tasks. Future work involves extending our framework to additional LMMs and developing methods for synthesizing high-quality multimodal reasoning data for rule-based reinforcement learning.

{
    \small
    \bibliographystyle{ieeenat_fullname}
    \bibliography{main}

\begin{thebibliography}{97}
\providecommand{\natexlab}[1]{#1}
\providecommand{\url}[1]{\texttt{#1}}
\expandafter\ifx\csname urlstyle\endcsname\relax
  \providecommand{\doi}[1]{doi: #1}\else
  \providecommand{\doi}{doi: \begingroup \urlstyle{rm}\Url}\fi

\bibitem[Achiam et~al.(2023)Achiam, Adler, Agarwal, Ahmad, Akkaya, Aleman, Almeida, Altenschmidt, Altman, Anadkat, et~al.]{gpt4}
Josh Achiam, Steven Adler, Sandhini Agarwal, Lama Ahmad, Ilge Akkaya, Florencia~Leoni Aleman, Diogo Almeida, Janko Altenschmidt, Sam Altman, Shyamal Anadkat, et~al.
\newblock {GPT-4} technical report.
\newblock \emph{arXiv preprint arXiv:2303.08774}, 2023.

\bibitem[Agrawal et~al.(2015)Agrawal, Lu, Antol, Mitchell, Zitnick, Parikh, and Batra]{vqa-as}
Aishwarya Agrawal, Jiasen Lu, Stanislaw Antol, Margaret Mitchell, C.~Lawrence Zitnick, Devi Parikh, and Dhruv Batra.
\newblock Vqa: Visual question answering.
\newblock \emph{IJCV}, 2015.

\bibitem[Alayrac et~al.(2022)Alayrac, Donahue, Luc, Miech, Barr, Hasson, Lenc, Mensch, Millican, Reynolds, et~al.]{flamingo}
Jean-Baptiste Alayrac, Jeff Donahue, Pauline Luc, Antoine Miech, Iain Barr, Yana Hasson, Karel Lenc, Arthur Mensch, Katherine Millican, Malcolm Reynolds, et~al.
\newblock Flamingo: A visual language model for few-shot learning.
\newblock In \emph{NeurIPS}, 2022.

\bibitem[Bai et~al.(2023{\natexlab{a}})Bai, Bai, Chu, Cui, Dang, Deng, Fan, Ge, Han, Huang, Hui, Ji, Li, Lin, Lin, Liu, Liu, Lu, Lu, Ma, Men, Ren, Ren, Tan, Tan, Tu, Wang, Wang, Wang, Wu, Xu, Xu, Yang, Yang, Yang, Yang, Yao, Yu, Yuan, Yuan, Zhang, Zhang, Zhang, Zhang, Zhou, Zhou, Zhou, and Zhu]{qwen}
Jinze Bai, Shuai Bai, Yunfei Chu, Zeyu Cui, Kai Dang, Xiaodong Deng, Yang Fan, Wenbin Ge, Yu Han, Fei Huang, Binyuan Hui, Luo Ji, Mei Li, Junyang Lin, Runji Lin, Dayiheng Liu, Gao Liu, Chengqiang Lu, Keming Lu, Jianxin Ma, Rui Men, Xingzhang Ren, Xuancheng Ren, Chuanqi Tan, Sinan Tan, Jianhong Tu, Peng Wang, Shijie Wang, Wei Wang, Shengguang Wu, Benfeng Xu, Jin Xu, An Yang, Hao Yang, Jian Yang, Shusheng Yang, Yang Yao, Bowen Yu, Hongyi Yuan, Zheng Yuan, Jianwei Zhang, Xingxuan Zhang, Yichang Zhang, Zhenru Zhang, Chang Zhou, Jingren Zhou, Xiaohuan Zhou, and Tianhang Zhu.
\newblock Qwen technical report.
\newblock \emph{arXiv preprint arXiv:2309.16609}, 2023{\natexlab{a}}.

\bibitem[Bai et~al.(2023{\natexlab{b}})Bai, Bai, Yang, Wang, Tan, Wang, Lin, Zhou, and Zhou]{qwenvl}
Jinze Bai, Shuai Bai, Shusheng Yang, Shijie Wang, Sinan Tan, Peng Wang, Junyang Lin, Chang Zhou, and Jingren Zhou.
\newblock {Qwen-VL}: A versatile vision-language model for understanding, localization, text reading, and beyond.
\newblock \emph{arXiv preprint arXiv:2308.12966}, 2023{\natexlab{b}}.

\bibitem[Bai et~al.(2025)Bai, Chen, Liu, Wang, Ge, Song, Dang, Wang, Wang, Tang, Zhong, Zhu, Yang, Li, Wan, Wang, Ding, Fu, Xu, Ye, Zhang, Xie, Cheng, Zhang, Yang, Xu, and Lin]{qwen2.5vl}
Shuai Bai, Keqin Chen, Xuejing Liu, Jialin Wang, Wenbin Ge, Sibo Song, Kai Dang, Peng Wang, Shijie Wang, Jun Tang, Humen Zhong, Yuanzhi Zhu, Mingkun Yang, Zhaohai Li, Jianqiang Wan, Pengfei Wang, Wei Ding, Zheren Fu, Yiheng Xu, Jiabo Ye, Xi Zhang, Tianbao Xie, Zesen Cheng, Hang Zhang, Zhibo Yang, Haiyang Xu, and Junyang Lin.
\newblock {Qwen2.5-VL} technical report.
\newblock \emph{arXiv preprint arXiv:2502.13923}, 2025.

\bibitem[Bai et~al.(2022)Bai, Jones, Ndousse, Askell, Chen, DasSarma, Drain, Fort, Ganguli, Henighan, et~al.]{bai2022training}
Yuntao Bai, Andy Jones, Kamal Ndousse, Amanda Askell, Anna Chen, Nova DasSarma, Dawn Drain, Stanislav Fort, Deep Ganguli, Tom Henighan, et~al.
\newblock Training a helpful and harmless assistant with reinforcement learning from human feedback.
\newblock \emph{arXiv preprint arXiv:2204.05862}, 2022.

\bibitem[Cao and Xiao(2022)]{geoqa+}
Jie Cao and Jing Xiao.
\newblock An augmented benchmark dataset for geometric question answering through dual parallel text encoding.
\newblock In \emph{International Conference on Computational Linguistics}, 2022.

\bibitem[Chang et~al.(2022)Chang, Palzer, Li, Fosler-Lussier, and Xiao]{mapqa}
Shuaichen Chang, David Palzer, Jialin Li, Eric Fosler-Lussier, and Ningchuan Xiao.
\newblock Mapqa: A dataset for question answering on choropleth maps.
\newblock \emph{arXiv preprint arXiv:2211.08545}, 2022.

\bibitem[Chen et~al.(2022)Chen, Li, Qin, Lu, Lin, Chen, and Liang]{unigeo}
Jiaqi Chen, Tong Li, Jinghui Qin, Pan Lu, Liang Lin, Chongyu Chen, and Xiaodan Liang.
\newblock {UniGeo}: Unifying geometry logical reasoning via reformulating mathematical expression.
\newblock \emph{arXiv preprint arXiv:2212.02746}, 2022.

\bibitem[Chen et~al.(2023)Chen, Zhang, Zeng, Zhang, Zhu, and Zhao]{shikra}
Keqin Chen, Zhao Zhang, Weili Zeng, Richong Zhang, Feng Zhu, and Rui Zhao.
\newblock Shikra: Unleashing multimodal llm's referential dialogue magic.
\newblock \emph{arXiv preprint arXiv:2306.15195}, 2023.

\bibitem[Chen et~al.(2024{\natexlab{a}})Chen, Li, Dong, Zhang, Zang, Chen, Duan, Wang, Qiao, Lin, and Zhao]{mmstar}
Lin Chen, Jinsong Li, Xiaoyi Dong, Pan Zhang, Yuhang Zang, Zehui Chen, Haodong Duan, Jiaqi Wang, Yu Qiao, Dahua Lin, and Feng Zhao.
\newblock Are we on the right way for evaluating large vision-language models?
\newblock In \emph{NeurIPS}, 2024{\natexlab{a}}.

\bibitem[Chen et~al.(2025{\natexlab{a}})Chen, Li, Zhao, Song, and Vinci]{chen2025r1v}
Liang Chen, Lei Li, Haozhe Zhao, Yifan Song, and Vinci.
\newblock R1-v: Reinforcing super generalization ability in vision-language models with less than \$3.
\newblock \url{https://github.com/Deep-Agent/R1-V}, 2025{\natexlab{a}}.
\newblock Accessed: 2025-02-02.

\bibitem[Chen et~al.(2025{\natexlab{b}})Chen, Wu, Liu, Pan, Liu, Xie, Yu, and Ruan]{januspro}
Xiaokang Chen, Zhiyu Wu, Xingchao Liu, Zizheng Pan, Wen Liu, Zhenda Xie, Xingkai Yu, and Chong Ruan.
\newblock Janus-pro: Unified multimodal understanding and generation with data and model scaling.
\newblock \emph{arXiv preprint arXiv:2501.17811}, 2025{\natexlab{b}}.

\bibitem[Chen et~al.(2024{\natexlab{b}})Chen, Wu, Wang, Su, Chen, Xing, Zhong, Zhang, Zhu, Lu, et~al.]{internvl}
Zhe Chen, Jiannan Wu, Wenhai Wang, Weijie Su, Guo Chen, Sen Xing, Muyan Zhong, Qinglong Zhang, Xizhou Zhu, Lewei Lu, et~al.
\newblock {InternVL}: Scaling up vision foundation models and aligning for generic visual-linguistic tasks.
\newblock In \emph{CVPR}, 2024{\natexlab{b}}.

\bibitem[Cheng et~al.(2024)Cheng, Chen, Zhang, Fei, Feng, Che, Li, and Qin]{comt}
Zihui Cheng, Qiguang Chen, Jin Zhang, Hao Fei, Xiaocheng Feng, Wanxiang Che, Min Li, and Libo Qin.
\newblock {CoMT}: A novel benchmark for chain of multi-modal thought on large vision-language models.
\newblock \emph{arXiv preprint arXiv:2412.12932}, 2024.

\bibitem[Chiang et~al.(2023)Chiang, Li, Lin, Sheng, Wu, Zhang, Zheng, Zhuang, Zhuang, Gonzalez, Stoica, and Xing]{vicuna}
Wei-Lin Chiang, Zhuohan Li, Zi Lin, Ying Sheng, Zhanghao Wu, Hao Zhang, Lianmin Zheng, Siyuan Zhuang, Yonghao Zhuang, Joseph~E. Gonzalez, Ion Stoica, and Eric~P. Xing.
\newblock Vicuna: An open-source chatbot impressing gpt-4 with 90\%* chatgpt quality, 2023.

\bibitem[Fourrier et~al.(2023)Fourrier, Habib, Kydlíček, Wolf, and Tunstall]{lighteval}
Clémentine Fourrier, Nathan Habib, Hynek Kydlíček, Thomas Wolf, and Lewis Tunstall.
\newblock {LightEval}: A lightweight framework for llm evaluation, 2023.

\bibitem[Gao et~al.(2024)Gao, Song, Yang, Cai, Miao, Dong, Li, Ma, Chen, Xu, Tang, Wang, Zan, Quan, Zhang, Sha, Zhang, Ren, Liu, and Chang]{omni-math}
Bofei Gao, Feifan Song, Zhe Yang, Zefan Cai, Yibo Miao, Qingxiu Dong, Lei Li, Chenghao Ma, Liang Chen, Runxin Xu, Zhengyang Tang, Benyou Wang, Daoguang Zan, Shanghaoran Quan, Ge Zhang, Lei Sha, Yichang Zhang, Xuancheng Ren, Tianyu Liu, and Baobao Chang.
\newblock {Omni-MATH}: A universal olympiad level mathematic benchmark for large language models.
\newblock \emph{arXiv preprint arXiv:2410.07985}, 2024.

\bibitem[Ge et~al.(2024)Ge, Zhao, Zhu, Ge, Yi, Song, Li, Ding, and Shan]{seedx}
Yuying Ge, Sijie Zhao, Jinguo Zhu, Yixiao Ge, Kun Yi, Lin Song, Chen Li, Xiaohan Ding, and Ying Shan.
\newblock Seed-x: Multimodal models with unified multi-granularity comprehension and generation.
\newblock \emph{arXiv preprint arXiv:2404.14396}, 2024.

\bibitem[Goyal et~al.(2016)Goyal, Khot, Summers-Stay, Batra, and Parikh]{vqa2.0}
Yash Goyal, Tejas Khot, Douglas Summers-Stay, Dhruv Batra, and Devi Parikh.
\newblock Making the v in vqa matter: Elevating the role of image understanding in visual question answering.
\newblock \emph{IJCV}, 2016.

\bibitem[Guo et~al.(2025)Guo, Yang, Zhang, Song, Zhang, Xu, Zhu, Ma, Wang, Bi, et~al.]{deepseekr1}
Daya Guo, Dejian Yang, Haowei Zhang, Junxiao Song, Ruoyu Zhang, Runxin Xu, Qihao Zhu, Shirong Ma, Peiyi Wang, Xiao Bi, et~al.
\newblock Deepseek-r1: Incentivizing reasoning capability in llms via reinforcement learning.
\newblock \emph{arXiv preprint arXiv:2501.12948}, 2025.

\bibitem[Gurari et~al.(2018)Gurari, Li, Stangl, Guo, Lin, Grauman, Luo, and Bigham]{vizwiz}
Danna Gurari, Qing Li, Abigale Stangl, Anhong Guo, Chi Lin, Kristen Grauman, Jiebo Luo, and Jeffrey~P. Bigham.
\newblock Vizwiz grand challenge: Answering visual questions from blind people.
\newblock In \emph{CVPR}, 2018.

\bibitem[He et~al.(2024)He, Luo, Bai, Hu, Thai, Shen, Hu, Han, Huang, Zhang, Liu, Qi, Liu, and Sun]{olympiadbench}
Chaoqun He, Renjie Luo, Yuzhuo Bai, Shengding Hu, Zhen~Leng Thai, Junhao Shen, Jinyi Hu, Xu Han, Yujie Huang, Yuxiang Zhang, Jie Liu, Lei Qi, Zhiyuan Liu, and Maosong Sun.
\newblock {OlympiadBench}: A challenging benchmark for promoting agi with olympiad-level bilingual multimodal scientific problems.
\newblock In \emph{ACL}, 2024.

\bibitem[Hu et~al.(2024)Hu, Wu, Zhu, Xianyu, Wang, Zhang, and Cao]{openrlhf}
Jian Hu, Xibin Wu, Zilin Zhu, Xianyu, Weixun Wang, Dehao Zhang, and Yu Cao.
\newblock {OpenRLHF}: An easy-to-use, scalable and high-performance {RLHF} framework.
\newblock \emph{arXiv preprint arXiv:2405.11143}, 2024.

\bibitem[Kafle et~al.(2018)Kafle, Cohen, Price, and Kanan]{dvqa}
Kushal Kafle, Scott~D. Cohen, Brian~L. Price, and Christopher Kanan.
\newblock {DVQA}: Understanding data visualizations via question answering.
\newblock \emph{CVPR}, 2018.

\bibitem[Kahou et~al.(2017)Kahou, Atkinson, Michalski, K{\'a}d{\'a}r, Trischler, and Bengio]{figureqa}
Samira~Ebrahimi Kahou, Adam Atkinson, Vincent Michalski, {\'A}kos K{\'a}d{\'a}r, Adam Trischler, and Yoshua Bengio.
\newblock Figureqa: An annotated figure dataset for visual reasoning.
\newblock \emph{arXiv preprint arXiv:1710.07300}, 2017.

\bibitem[Kembhavi et~al.(2016)Kembhavi, Salvato, Kolve, Seo, Hajishirzi, and Farhadi]{ai2d}
Aniruddha Kembhavi, Michael Salvato, Eric Kolve, Minjoon Seo, Hannaneh Hajishirzi, and Ali Farhadi.
\newblock A diagram is worth a dozen images.
\newblock \emph{arXiv preprint arXiv:1603.07396}, 2016.

\bibitem[Kembhavi et~al.(2017)Kembhavi, Seo, Schwenk, Choi, Farhadi, and Hajishirzi]{tqa}
Aniruddha Kembhavi, Minjoon Seo, Dustin Schwenk, Jonghyun Choi, Ali Farhadi, and Hannaneh Hajishirzi.
\newblock Are you smarter than a sixth grader? textbook question answering for multimodal machine comprehension.
\newblock \emph{CVPR}, 2017.

\bibitem[Kurach et~al.(2020)Kurach, Raichuk, Sta{\'n}czyk, Zaj{\k{a}}c, Bachem, Espeholt, Riquelme, Vincent, Michalski, Bousquet, et~al.]{football}
Karol Kurach, Anton Raichuk, Piotr Sta{\'n}czyk, Micha{\l} Zaj{\k{a}}c, Olivier Bachem, Lasse Espeholt, Carlos Riquelme, Damien Vincent, Marcin Michalski, Olivier Bousquet, et~al.
\newblock Google research football: A novel reinforcement learning environment.
\newblock In \emph{AAAI}, 2020.

\bibitem[Kwon et~al.(2023)Kwon, Li, Zhuang, Sheng, Zheng, Yu, Gonzalez, Zhang, and Stoica]{vllm}
Woosuk Kwon, Zhuohan Li, Siyuan Zhuang, Ying Sheng, Lianmin Zheng, Cody~Hao Yu, Joseph~E. Gonzalez, Hao Zhang, and Ion Stoica.
\newblock Efficient memory management for large language model serving with pagedattention.
\newblock In \emph{Proceedings of the ACM SIGOPS 29th Symposium on Operating Systems Principles}, 2023.

\bibitem[Lau et~al.(2018)Lau, Gayen, Abacha, and Demner-Fushman]{vqarad}
Jason~J Lau, Soumya Gayen, Asma~Ben Abacha, and Dina Demner-Fushman.
\newblock A dataset of clinically generated visual questions and answers about radiology images.
\newblock \emph{Scientific Data}, 2018.

\bibitem[Li et~al.(2023)Li, Li, Savarese, and Hoi]{blip2}
Junnan Li, Dongxu Li, Silvio Savarese, and Steven Hoi.
\newblock {BLIP-2}: Bootstrapping language-image pre-training with frozen image encoders and large language models.
\newblock In \emph{ICML}, 2023.

\bibitem[Li et~al.(2025)Li, Jiang, Hu, Wang, Zhong, Luo, Ma, and Zhang]{unimoe}
Yunxin Li, Shenyuan Jiang, Baotian Hu, Longyue Wang, Wanqi Zhong, Wenhan Luo, Lin Ma, and Min Zhang.
\newblock {Uni-mMoE}: Scaling unified multimodal llms with mixture of experts.
\newblock \emph{IEEE Transactions on Pattern Analysis and Machine Intelligence}, 2025.

\bibitem[Li et~al.(2022)Li, Wang, Stengel-Eskin, Kortylewski, Ma, Durme, University, of~Southern~California, for Informatics, and of~Freiburg]{super-clevr}
Zhuowan Li, Xingrui Wang, Elias Stengel-Eskin, Adam Kortylewski, Wufei Ma, Benjamin~Van Durme, Alan Yuille Johns~Hopkins University, University of Southern~California, Max Planck~Institute for Informatics, and University of Freiburg.
\newblock {Super-CLEVR}: A virtual benchmark to diagnose domain robustness in visual reasoning.
\newblock In \emph{CVPR}, 2022.

\bibitem[Lightman et~al.(2023)Lightman, Kosaraju, Burda, Edwards, Baker, Lee, Leike, Schulman, Sutskever, and Cobbe]{stepbystep}
Hunter Lightman, Vineet Kosaraju, Yuri Burda, Harrison Edwards, Bowen Baker, Teddy Lee, Jan Leike, John Schulman, Ilya Sutskever, and Karl Cobbe.
\newblock Let's verify step by step.
\newblock In \emph{ICLR}, 2023.

\bibitem[Lin et~al.(2024)Lin, Yin, Ping, Molchanov, Shoeybi, and Han]{vila}
Ji Lin, Hongxu Yin, Wei Ping, Pavlo Molchanov, Mohammad Shoeybi, and Song Han.
\newblock {VILA}: On pre-training for visual language models.
\newblock In \emph{CVPR}, 2024.

\bibitem[Lindstr{\"o}m and Abraham(2022)]{clevr-math}
Adam~Dahlgren Lindstr{\"o}m and Savitha~Sam Abraham.
\newblock {CLEVR-Math}: A dataset for compositional language, visual and mathematical reasoning.
\newblock \emph{arXiv preprint arXiv:2208.05358}, 2022.

\bibitem[Liu et~al.(2023)Liu, Li, Wu, and Lee]{llava}
Haotian Liu, Chunyuan Li, Qingyang Wu, and Yong~Jae Lee.
\newblock Visual instruction tuning.
\newblock In \emph{NeurIPS}, 2023.

\bibitem[Liu et~al.(2024{\natexlab{a}})Liu, Li, Li, and Lee]{llava1.5}
Haotian Liu, Chunyuan Li, Yuheng Li, and Yong~Jae Lee.
\newblock Improved baselines with visual instruction tuning.
\newblock In \emph{CVPR}, 2024{\natexlab{a}}.

\bibitem[Liu et~al.(2024{\natexlab{b}})Liu, Li, Li, Li, Zhang, Shen, and Lee]{llavanext}
Haotian Liu, Chunyuan Li, Yuheng Li, Bo Li, Yuanhan Zhang, Sheng Shen, and Yong~Jae Lee.
\newblock {LLaVA-NeXT}: Improved reasoning, ocr, and world knowledge, 2024{\natexlab{b}}.

\bibitem[Liu et~al.(2025)Liu, Zhang, Gu, Iong, XiXuan, Xu, Zhang, Lai, Sun, Yang, et~al.]{visualagentbench}
Xiao Liu, Tianjie Zhang, Yu Gu, Iat~Long Iong, Song XiXuan, Yifan Xu, Shudan Zhang, Hanyu Lai, Jiadai Sun, Xinyue Yang, et~al.
\newblock Visualagentbench: Towards large multimodal models as visual agents.
\newblock In \emph{ICLR}, 2025.

\bibitem[Lu et~al.(2021{\natexlab{a}})Lu, Gong, Jiang, Qiu, Huang, Liang, and Zhu]{geometry3k}
Pan Lu, Ran Gong, Shibiao Jiang, Liang Qiu, Siyuan Huang, Xiaodan Liang, and Song-Chun Zhu.
\newblock {Inter-GPS}: Interpretable geometry problem solving with formal language and symbolic reasoning.
\newblock In \emph{ACL}, 2021{\natexlab{a}}.

\bibitem[Lu et~al.(2021{\natexlab{b}})Lu, Qiu, Chen, Xia, Zhao, Zhang, Yu, Liang, and Zhu]{iconqa}
Pan Lu, Liang Qiu, Jiaqi Chen, Tony Xia, Yizhou Zhao, Wei Zhang, Zhou Yu, Xiaodan Liang, and Song-Chun Zhu.
\newblock {IconQA}: A new benchmark for abstract diagram understanding and visual language reasoning.
\newblock \emph{arXiv preprint arXiv:2110.13214}, 2021{\natexlab{b}}.

\bibitem[Lu et~al.(2022{\natexlab{a}})Lu, Mishra, Xia, Qiu, Chang, Zhu, Tafjord, Clark, and Kalyan]{scienceqa}
Pan Lu, Swaroop Mishra, Tony Xia, Liang Qiu, Kai-Wei Chang, Song-Chun Zhu, Oyvind Tafjord, Peter Clark, and A. Kalyan.
\newblock Learn to explain: Multimodal reasoning via thought chains for science question answering.
\newblock \emph{arXiv preprint arXiv:2209.09513}, 2022{\natexlab{a}}.

\bibitem[Lu et~al.(2022{\natexlab{b}})Lu, Qiu, Chang, Wu, Zhu, Rajpurohit, Clark, and Kalyan]{tabmwp}
Pan Lu, Liang Qiu, Kai-Wei Chang, Ying~Nian Wu, Song-Chun Zhu, Tanmay Rajpurohit, Peter Clark, and A. Kalyan.
\newblock Dynamic prompt learning via policy gradient for semi-structured mathematical reasoning.
\newblock \emph{arXiv preprint arXiv:2209.14610}, 2022{\natexlab{b}}.

\bibitem[Lu et~al.(2024)Lu, Bansal, Xia, Liu, Li, Hajishirzi, Cheng, Chang, Galley, and Gao]{mathvista}
Pan Lu, Hritik Bansal, Tony Xia, Jiacheng Liu, Chunyuan Li, Hannaneh Hajishirzi, Hao Cheng, Kai-Wei Chang, Michel Galley, and Jianfeng Gao.
\newblock {MathVista}: Evaluating mathematical reasoning of foundation models in visual contexts.
\newblock In \emph{ICLR}, 2024.

\bibitem[Luo et~al.(2024)Luo, Liu, Liu, Phatale, Lara, Li, Shu, Zhu, Meng, Sun, et~al.]{omega_prm}
Liangchen Luo, Yinxiao Liu, Rosanne Liu, Samrat Phatale, Harsh Lara, Yunxuan Li, Lei Shu, Yun Zhu, Lei Meng, Jiao Sun, et~al.
\newblock Improve mathematical reasoning in language models by automated process supervision.
\newblock \emph{arXiv preprint arXiv:2406.06592}, 2, 2024.

\bibitem[Luo et~al.(2025{\natexlab{a}})Luo, Tan, Wong, Shi, Tang, Roongta, Cai, Luo, Zhang, Li, Popa, and Stoica]{deepscaler}
Michael Luo, Sijun Tan, Justin Wong, Xiaoxiang Shi, William~Y. Tang, Manan Roongta, Colin Cai, Jeffrey Luo, Tianjun Zhang, Li~Erran Li, Raluca~Ada Popa, and Ion Stoica.
\newblock {DeepScaleR}: Surpassing o1-preview with a 1.5b model by scaling rl, 2025{\natexlab{a}}.
\newblock Notion Blog.

\bibitem[Luo et~al.(2025{\natexlab{b}})Luo, Zheng, Wang, Yu, Ni, Lin, Zeng, and Yang]{ursa}
Ruilin Luo, Zhuofan Zheng, Yifan Wang, Yiyao Yu, Xinzhe Ni, Zicheng Lin, Jin Zeng, and Yujiu Yang.
\newblock Ursa: Understanding and verifying chain-of-thought reasoning in multimodal mathematics.
\newblock \emph{arXiv preprint arXiv:2501.04686}, 2025{\natexlab{b}}.

\bibitem[Masry et~al.(2022)Masry, Long, Tan, Joty, and Hoque]{chartqa}
Ahmed Masry, Do~Xuan Long, Jia~Qing Tan, Shafiq~R. Joty, and Enamul Hoque.
\newblock Chartqa: A benchmark for question answering about charts with visual and logical reasoning.
\newblock \emph{arXiv preprint arXiv:2203.10244}, 2022.

\bibitem[Mathew et~al.(2021)Mathew, Bagal, Tito, Karatzas, Valveny, and Jawahar]{docvqa}
Minesh Mathew, Viraj Bagal, Rub{\`e}n~P{\'e}rez Tito, Dimosthenis Karatzas, Ernest Valveny, and C.V. Jawahar.
\newblock Infographicvqa.
\newblock In \emph{WACV}, 2021.

\bibitem[Methani et~al.(2019)Methani, Ganguly, Khapra, and Kumar]{plotqa}
Nitesh Methani, Pritha Ganguly, Mitesh~M. Khapra, and Pratyush Kumar.
\newblock Plotqa: Reasoning over scientific plots.
\newblock In \emph{WACV}, 2019.

\bibitem[OpenAI(2023)]{gpt4v}
OpenAI.
\newblock Gpt-4v(ision) system card, 2023.

\bibitem[OpenAI(2024)]{o1}
OpenAI.
\newblock Openai o1 system card, 2024.

\bibitem[Ouyang et~al.(2022)Ouyang, Wu, Jiang, Almeida, Wainwright, Mishkin, Zhang, Agarwal, Slama, Ray, et~al.]{ouyang2022training}
Long Ouyang, Jeffrey Wu, Xu Jiang, Diogo Almeida, Carroll Wainwright, Pamela Mishkin, Chong Zhang, Sandhini Agarwal, Katarina Slama, Alex Ray, et~al.
\newblock Training language models to follow instructions with human feedback.
\newblock In \emph{NeurIPS}, 2022.

\bibitem[Peng et~al.(2023)Peng, Wang, Dong, Hao, Huang, Ma, and Wei]{kosmos2}
Zhiliang Peng, Wenhui Wang, Li Dong, Yaru Hao, Shaohan Huang, Shuming Ma, and Furu Wei.
\newblock Kosmos-2: Grounding multimodal large language models to the world.
\newblock \emph{arXiv preprint arXiv:2306.14824}, 2023.

\bibitem[Racani{\`e}re et~al.(2017)Racani{\`e}re, Weber, Reichert, Buesing, Guez, Jimenez~Rezende, Puigdom{\`e}nech~Badia, Vinyals, Heess, Li, et~al.]{sokoban}
S{\'e}bastien Racani{\`e}re, Th{\'e}ophane Weber, David Reichert, Lars Buesing, Arthur Guez, Danilo Jimenez~Rezende, Adri{\`a} Puigdom{\`e}nech~Badia, Oriol Vinyals, Nicolas Heess, Yujia Li, et~al.
\newblock Imagination-augmented agents for deep reinforcement learning.
\newblock In \emph{NeurIPS}, 2017.

\bibitem[Rein et~al.(2024)Rein, Hou, Stickland, Petty, Pang, Dirani, Michael, and Bowman]{gpqa}
David Rein, Betty~Li Hou, Asa~Cooper Stickland, Jackson Petty, Richard~Yuanzhe Pang, Julien Dirani, Julian Michael, and Samuel~R Bowman.
\newblock {GPQA}: A graduate-level google-proof q\&a benchmark.
\newblock In \emph{COLM}, 2024.

\bibitem[Schulman et~al.(2017)Schulman, Wolski, Dhariwal, Radford, and Klimov]{ppo}
John Schulman, Filip Wolski, Prafulla Dhariwal, Alec Radford, and Oleg Klimov.
\newblock Proximal policy optimization algorithms.
\newblock \emph{arXiv preprint arXiv:1707.06347}, 2017.

\bibitem[Seo et~al.(2015)Seo, Hajishirzi, Farhadi, Etzioni, and Malcolm]{geos}
Minjoon Seo, Hannaneh Hajishirzi, Ali Farhadi, Oren Etzioni, and Clint Malcolm.
\newblock Solving geometry problems: Combining text and diagram interpretation.
\newblock In \emph{EMNLP}, 2015.

\bibitem[Setlur et~al.(2025)Setlur, Nagpal, Fisch, Geng, Eisenstein, Agarwal, Agarwal, Berant, and Kumar]{pav}
Amrith Setlur, Chirag Nagpal, Adam Fisch, Xinyang Geng, Jacob Eisenstein, Rishabh Agarwal, Alekh Agarwal, Jonathan Berant, and Aviral Kumar.
\newblock Rewarding progress: Scaling automated process verifiers for llm reasoning.
\newblock In \emph{ICLR}, 2025.

\bibitem[Shen et~al.(2025)Shen, Chen, Shao, Guan, and Nie]{mome}
Leyang Shen, Gongwei Chen, Rui Shao, Weili Guan, and Liqiang Nie.
\newblock {MoME}: Mixture of multimodal experts for generalist multimodal large language models.
\newblock In \emph{NeurIPS}, 2025.

\bibitem[Shi et~al.(2024)Shi, Hu, Bin, Liu, Yang, Ng, Bing, and wei Lee]{mathv}
Wenhao Shi, Zhiqiang Hu, Yi Bin, Junhua Liu, Yang Yang, See-Kiong Ng, Li Bing, and Roy~Ka wei Lee.
\newblock {Math-LLaVA}: Bootstrapping mathematical reasoning for multimodal large language models.
\newblock In \emph{EMNLP}, 2024.

\bibitem[Singh et~al.(2019)Singh, Natarajan, Shah, Jiang, Chen, Batra, Parikh, and Rohrbach]{textvqa}
Amanpreet Singh, Vivek Natarajan, Meet Shah, Yu Jiang, Xinlei Chen, Dhruv Batra, Devi Parikh, and Marcus Rohrbach.
\newblock Towards vqa models that can read.
\newblock In \emph{CVPR}, 2019.

\bibitem[Sun et~al.(2023)Sun, Shen, Cao, Liu, Li, Shen, Gan, Gui, Wang, Yang, et~al.]{sun2023aligning}
Zhiqing Sun, Sheng Shen, Shengcao Cao, Haotian Liu, Chunyuan Li, Yikang Shen, Chuang Gan, Liang-Yan Gui, Yu-Xiong Wang, Yiming Yang, et~al.
\newblock Aligning large multimodal models with factually augmented {RLHF}.
\newblock \emph{arXiv preprint arXiv:2309.14525}, 2023.

\bibitem[Team(2024{\natexlab{a}})]{chameleon}
Chameleon Team.
\newblock Chameleon: Mixed-modal early-fusion foundation models.
\newblock \emph{arXiv preprint arXiv:2405.09818}, 2024{\natexlab{a}}.

\bibitem[Team et~al.(2023)Team, Anil, Borgeaud, Alayrac, Yu, Soricut, Schalkwyk, Dai, Hauth, Millican, et~al.]{gemini}
Gemini Team, Rohan Anil, Sebastian Borgeaud, Jean-Baptiste Alayrac, Jiahui Yu, Radu Soricut, Johan Schalkwyk, Andrew~M Dai, Anja Hauth, Katie Millican, et~al.
\newblock {Gemini}: a family of highly capable multimodal models.
\newblock \emph{arXiv preprint arXiv:2312.11805}, 2023.

\bibitem[Team et~al.(2025)Team, Du, Gao, Xing, Jiang, Chen, Li, Xiao, Du, Liao, et~al.]{team2025kimi}
Kimi Team, Angang Du, Bofei Gao, Bowei Xing, Changjiu Jiang, Cheng Chen, Cheng Li, Chenjun Xiao, Chenzhuang Du, Chonghua Liao, et~al.
\newblock Kimi k1. 5: Scaling reinforcement learning with llms.
\newblock \emph{arXiv preprint arXiv:2501.12599}, 2025.

\bibitem[Team(2024{\natexlab{b}})]{qwen2.5}
Qwen Team.
\newblock Qwen2.5: A party of foundation models, 2024{\natexlab{b}}.

\bibitem[Team(2024{\natexlab{c}})]{qwq}
Qwen Team.
\newblock Qwq: Reflect deeply on the boundaries of the unknown, 2024{\natexlab{c}}.

\bibitem[Team(2025)]{still}
RUCAIBox~STILL Team.
\newblock {STILL-3-1.5B-preview}: Enhancing slow thinking abilities of small models through reinforcement learning, 2025.

\bibitem[Touvron et~al.(2023)Touvron, Lavril, Izacard, Martinet, Lachaux, Lacroix, Rozi{\`e}re, Goyal, Hambro, Azhar, et~al.]{llama}
Hugo Touvron, Thibaut Lavril, Gautier Izacard, Xavier Martinet, Marie-Anne Lachaux, Timoth{\'e}e Lacroix, Baptiste Rozi{\`e}re, Naman Goyal, Eric Hambro, Faisal Azhar, et~al.
\newblock Llama: Open and efficient foundation language models.
\newblock \emph{arXiv preprint arXiv:2302.13971}, 2023.

\bibitem[Wang et~al.(2025{\natexlab{a}})Wang, Xu, Jia, Zhang, Yan, Shen, Zhang, Huang, and Sang]{mobileagentv2}
Junyang Wang, Haiyang Xu, Haitao Jia, Xi Zhang, Ming Yan, Weizhou Shen, Ji Zhang, Fei Huang, and Jitao Sang.
\newblock Mobile-agent-v2: Mobile device operation assistant with effective navigation via multi-agent collaboration.
\newblock In \emph{NeurIPS}, 2025{\natexlab{a}}.

\bibitem[Wang et~al.(2025{\natexlab{b}})Wang, Pan, Shi, Lu, Ren, Zhou, Zhan, and Li]{mathvision}
Ke Wang, Junting Pan, Weikang Shi, Zimu Lu, Houxing Ren, Aojun Zhou, Mingjie Zhan, and Hongsheng Li.
\newblock Measuring multimodal mathematical reasoning with math-vision dataset.
\newblock In \emph{NeurIPS}, 2025{\natexlab{b}}.

\bibitem[Wang et~al.(2023{\natexlab{a}})Wang, Li, Shao, Xu, Dai, Li, Chen, Wu, and Sui]{mathshepherd}
Peiyi Wang, Lei Li, Zhihong Shao, RX Xu, Damai Dai, Yifei Li, Deli Chen, Yu Wu, and Zhifang Sui.
\newblock Math-shepherd: Verify and reinforce llms step-by-step without human annotations.
\newblock \emph{arXiv preprint arXiv:2312.08935}, 2023{\natexlab{a}}.

\bibitem[Wang et~al.(2024{\natexlab{a}})Wang, Bai, Tan, Wang, Fan, Bai, Chen, Liu, Wang, Ge, Fan, Dang, Du, Ren, Men, Liu, Zhou, Zhou, and Lin]{qwen2vl}
Peng Wang, Shuai Bai, Sinan Tan, Shijie Wang, Zhihao Fan, Jinze Bai, Keqin Chen, Xuejing Liu, Jialin Wang, Wenbin Ge, Yang Fan, Kai Dang, Mengfei Du, Xuancheng Ren, Rui Men, Dayiheng Liu, Chang Zhou, Jingren Zhou, and Junyang Lin.
\newblock {Qwen2-VL}: Enhancing vision-language model's perception of the world at any resolution.
\newblock \emph{arXiv preprint arXiv:2409.12191}, 2024{\natexlab{a}}.

\bibitem[Wang et~al.(2023{\natexlab{b}})Wang, Chen, Chen, Wu, Zhu, Zeng, Luo, Lu, Zhou, Qiao, et~al.]{visionllm}
Wenhai Wang, Zhe Chen, Xiaokang Chen, Jiannan Wu, Xizhou Zhu, Gang Zeng, Ping Luo, Tong Lu, Jie Zhou, Yu Qiao, et~al.
\newblock {VisionLLM}: Large language model is also an open-ended decoder for vision-centric tasks.
\newblock In \emph{NeurIPS}, 2023{\natexlab{b}}.

\bibitem[Wang et~al.(2024{\natexlab{b}})Wang, Chen, Han, Lin, Zhao, Liu, Zhai, Yuan, You, and Yang]{lmm_reason_survey}
Yiqi Wang, Wentao Chen, Xiaotian Han, Xudong Lin, Haiteng Zhao, Yongfei Liu, Bohan Zhai, Jianbo Yuan, Quanzeng You, and Hongxia Yang.
\newblock Exploring the reasoning abilities of multimodal large language models (mllms): A comprehensive survey on emerging trends in multimodal reasoning.
\newblock \emph{arXiv preprint arXiv:2401.06805}, 2024{\natexlab{b}}.

\bibitem[Wu et~al.(2024{\natexlab{a}})Wu, Chen, Wu, Ma, Liu, Pan, Liu, Xie, Yu, Ruan, et~al.]{janus}
Chengyue Wu, Xiaokang Chen, Zhiyu Wu, Yiyang Ma, Xingchao Liu, Zizheng Pan, Wen Liu, Zhenda Xie, Xingkai Yu, Chong Ruan, et~al.
\newblock Janus: Decoupling visual encoding for unified multimodal understanding and generation.
\newblock \emph{arXiv preprint arXiv:2410.13848}, 2024{\natexlab{a}}.

\bibitem[Wu et~al.(2024{\natexlab{b}})Wu, Chen, Pan, Liu, Liu, Dai, Gao, Ma, Wu, Wang, et~al.]{deepseekvl2}
Zhiyu Wu, Xiaokang Chen, Zizheng Pan, Xingchao Liu, Wen Liu, Damai Dai, Huazuo Gao, Yiyang Ma, Chengyue Wu, Bingxuan Wang, et~al.
\newblock Deepseek-vl2: Mixture-of-experts vision-language models for advanced multimodal understanding.
\newblock \emph{arXiv preprint arXiv:2412.10302}, 2024{\natexlab{b}}.

\bibitem[Xiang et~al.(2024)Xiang, Liu, Jiang, Nie, Huang, Fan, Li, Huang, Zeng, Han, et~al.]{atomthink}
Kun Xiang, Zhili Liu, Zihao Jiang, Yunshuang Nie, Runhui Huang, Haoxiang Fan, Hanhui Li, Weiran Huang, Yihan Zeng, Jianhua Han, et~al.
\newblock Atomthink: A slow thinking framework for multimodal mathematical reasoning.
\newblock \emph{arXiv preprint arXiv:2411.11930}, 2024.

\bibitem[Xie et~al.(2024{\natexlab{a}})Xie, Mao, Bai, Zhang, Wang, Lin, Gu, Chen, Yang, and Shou]{showo}
Jinheng Xie, Weijia Mao, Zechen Bai, David~Junhao Zhang, Weihao Wang, Kevin~Qinghong Lin, Yuchao Gu, Zhijie Chen, Zhenheng Yang, and Mike~Zheng Shou.
\newblock Show-o: One single transformer to unify multimodal understanding and generation.
\newblock \emph{arXiv preprint arXiv:2408.12528}, 2024{\natexlab{a}}.

\bibitem[Xie et~al.(2024{\natexlab{b}})Xie, Goyal, Zheng, Kan, Lillicrap, Kawaguchi, and Shieh]{mcts_dpo}
Yuxi Xie, Anirudh Goyal, Wenyue Zheng, Min-Yen Kan, Timothy~P Lillicrap, Kenji Kawaguchi, and Michael Shieh.
\newblock Monte carlo tree search boosts reasoning via iterative preference learning.
\newblock \emph{arXiv preprint arXiv:2405.00451}, 2024{\natexlab{b}}.

\bibitem[Yao et~al.(2024)Yao, Huang, Wu, Zhang, Wang, Liu, Wang, Song, Feng, Shen, et~al.]{mulberry}
Huanjin Yao, Jiaxing Huang, Wenhao Wu, Jingyi Zhang, Yibo Wang, Shunyu Liu, Yingjie Wang, Yuxin Song, Haocheng Feng, Li Shen, et~al.
\newblock Mulberry: Empowering mllm with o1-like reasoning and reflection via collective monte carlo tree search.
\newblock \emph{arXiv preprint arXiv:2412.18319}, 2024.

\bibitem[Ye et~al.(2024{\natexlab{a}})Ye, Xu, Liu, Hu, Yan, Qian, Zhang, Huang, and Zhou]{mplugowl3}
Jiabo Ye, Haiyang Xu, Haowei Liu, Anwen Hu, Ming Yan, Qi Qian, Ji Zhang, Fei Huang, and Jingren Zhou.
\newblock {mPLUG-Owl3}: Towards long image-sequence understanding in multi-modal large language models.
\newblock In \emph{ICLR}, 2024{\natexlab{a}}.

\bibitem[Ye et~al.(2023)Ye, Xu, Xu, Ye, Yan, Zhou, Wang, Hu, Shi, Shi, et~al.]{mplugowl}
Qinghao Ye, Haiyang Xu, Guohai Xu, Jiabo Ye, Ming Yan, Yiyang Zhou, Junyang Wang, Anwen Hu, Pengcheng Shi, Yaya Shi, et~al.
\newblock {mPLUG-Owl}: Modularization empowers large language models with multimodality.
\newblock \emph{arXiv preprint arXiv:2304.14178}, 2023.

\bibitem[Ye et~al.(2024{\natexlab{b}})Ye, Xu, Ye, Yan, Hu, Liu, Qian, Zhang, and Huang]{mplugowl2}
Qinghao Ye, Haiyang Xu, Jiabo Ye, Ming Yan, Anwen Hu, Haowei Liu, Qi Qian, Ji Zhang, and Fei Huang.
\newblock {mPLUG-Owl2}: Revolutionizing multi-modal large language model with modality collaboration.
\newblock In \emph{CVPR}, 2024{\natexlab{b}}.

\bibitem[Zeng et~al.(2025)Zeng, Huang, Liu, He, Liu, Ma, and He]{simplerl-reasoning}
Weihao Zeng, Yuzhen Huang, Wei Liu, Keqing He, Qian Liu, Zejun Ma, and Junxian He.
\newblock 7b model and 8k examples: Emerging reasoning with reinforcement learning is both effective and efficient.
\newblock \url{https://hkust-nlp.notion.site/simplerl-reason}, 2025.
\newblock Notion Blog.

\bibitem[Zhai et~al.(2023)Zhai, Tong, Li, Cai, Qu, Lee, and Ma]{forgetting}
Yuexiang Zhai, Shengbang Tong, Xiao Li, Mu Cai, Qing Qu, Yong~Jae Lee, and Yi Ma.
\newblock Investigating the catastrophic forgetting in multimodal large language models.
\newblock In \emph{NeurIPS Workshops}, 2023.

\bibitem[Zhang et~al.(2024{\natexlab{a}})Zhang, Li, Zhang, Pu, Cahyono, Hu, Liu, Zhang, Yang, Li, and Liu]{lmmseval}
Kaichen Zhang, Bo Li, Peiyuan Zhang, Fanyi Pu, Joshua~Adrian Cahyono, Kairui Hu, Shuai Liu, Yuanhan Zhang, Jingkang Yang, Chunyuan Li, and Ziwei Liu.
\newblock Lmms-eval: Reality check on the evaluation of large multimodal models.
\newblock \emph{arXiv preprint arXiv:2407.12772}, 2024{\natexlab{a}}.

\bibitem[Zhang et~al.(2024{\natexlab{b}})Zhang, Dai, Yang, Bao, Chen, Qiu, Luo, Geng, and Guo]{magebench}
Miaosen Zhang, Qi Dai, Yifan Yang, Jianmin Bao, Dongdong Chen, Kai Qiu, Chong Luo, Xin Geng, and Baining Guo.
\newblock {MageBench}: Bridging large multimodal models to agents.
\newblock \emph{arXiv preprint arXiv:2412.04531}, 2024{\natexlab{b}}.

\bibitem[Zhang et~al.(2024{\natexlab{c}})Zhang, Jiang, Zhang, Lin, Guo, Qiu, Zhou, Lu, Chang, Qiao, Gao, and Li]{mathverse}
Renrui Zhang, Dongzhi Jiang, Yichi Zhang, Haokun Lin, Ziyu Guo, Pengshuo Qiu, Aojun Zhou, Pan Lu, Kai-Wei Chang, Yu Qiao, Peng Gao, and Hongsheng Li.
\newblock {MathVerse}: Does your multi-modal llm truly see the diagrams in visual math problems?
\newblock In \emph{ECCV}, 2024{\natexlab{c}}.

\bibitem[Zhang et~al.(2023)Zhang, Wu, Zhao, Lin, Zhang, Wang, and Xie]{pmcvqa}
Xiaoman Zhang, Chaoyi Wu, Ziheng Zhao, Weixiong Lin, Ya Zhang, Yanfeng Wang, and Weidi Xie.
\newblock {PMC-VQA}: Visual instruction tuning for medical visual question answering.
\newblock \emph{arXiv preprint arXiv:2305.10415}, 2023.

\bibitem[Zhou et~al.(2024)Zhou, Yu, Babu, Tirumala, Yasunaga, Shamis, Kahn, Ma, Zettlemoyer, and Levy]{transfusion}
Chunting Zhou, Lili Yu, Arun Babu, Kushal Tirumala, Michihiro Yasunaga, Leonid Shamis, Jacob Kahn, Xuezhe Ma, Luke Zettlemoyer, and Omer Levy.
\newblock Transfusion: Predict the next token and diffuse images with one multi-modal model.
\newblock \emph{arXiv preprint arXiv:2408.11039}, 2024.

\bibitem[Zhu et~al.(2024)Zhu, Chen, Shen, Li, and Elhoseiny]{minigpt4}
Deyao Zhu, Jun Chen, Xiaoqian Shen, Xiang Li, and Mohamed Elhoseiny.
\newblock {MiniGPT-4}: Enhancing vision-language understanding with advanced large language models.
\newblock In \emph{ICLR}, 2024.

\bibitem[Zong et~al.(2024)Zong, Ma, Shen, Song, Shao, Jiang, Li, and Liu]{mova}
Zhuofan Zong, Bingqi Ma, Dazhong Shen, Guanglu Song, Hao Shao, Dongzhi Jiang, Hongsheng Li, and Yu Liu.
\newblock {MoVA}: Adapting mixture of vision experts to multimodal context.
\newblock In \emph{NeurIPS}, 2024.

\end{thebibliography}
}

\clearpage

\renewcommand\thefigure{A\arabic{figure}}
\renewcommand\thetable{A\arabic{table}}  
\renewcommand\theequation{A\arabic{equation}}


\appendix
\section*{Appendix}

\section{Training Details}
\label{app:training_details}
\subsection{Datasets}
\label{app:train_datasets}

\noindent\textbf{Text-Only Dataset:} We use DeepScaleR-Preview~\cite{deepscaler} as the text-only math reasoning dataset to train FRE-Text. This dataset is composed of AIME(American Invitational Mathematics Examination) problems (1984-2023), AMC (American Mathematics Competition) problems (prior to 2023), Omni-MATH~\cite{omni-math} and STILL~\cite{still}. It is filtered to keep those that have verifiable answers and to remove redundant problems, leaving 40.3k high-quality data points.

\noindent\textbf{VerMulti-65K:} We use 65k verifiable problems filtered from MathV360K to train FRE-Multi, which consist of problems from various sources as shown in \cref{tab:mathv60k_statistics}.

\begin{table}[!h]
\centering
\setlength\tabcolsep{3pt}
\begin{tabular}{@{}lc@{}c@{}lc@{}}
\toprule
Split      & Count &\hspace{2em} &Split & Count\\ \midrule
IconQA~\cite{iconqa}       & 7166 &  & TQA~\cite{tqa}          & 2130 \\
PMC-VQA~\cite{pmcvqa}     & 6760 & & DocVQA~\cite{docvqa}       & 1974 \\
TabMWP~\cite{tabmwp} & 6732  &  &   TextVQA~\cite{textvqa}  & 1462 \\
A-OKVQA~\cite{ai2d}       & 6185  &  & VQA2.0~\cite{vqa2.0}       & 1316 \\
FigureQA~\cite{figureqa}     & 4995  & & ChartQA~\cite{chartqa}      & 1115 \\
ScienceQA~\cite{scienceqa}    & 4243  & & PlotQA~\cite{plotqa}       & 1020 \\
GeoQA+~\cite{geoqa+}       & 4062  & & Super-CLEVR~\cite{super-clevr}  & 1016 \\
DVQA~\cite{dvqa}         & 3317  & & VQA-AS~\cite{vqa-as}       & 404  \\
Geometry3K~\cite{geometry3k}   & 2845  & & MapQA~\cite{mapqa}        & 278  \\
UniGeo~\cite{unigeo}   & 2767  & & VizWiz~\cite{vizwiz}       & 204  \\
AI2D~\cite{ai2d}         & 2603  & & GEOS~\cite{geos}         & 129  \\
CLEVR-Math~\cite{clevr-math}   & 2393  & & VQA-RAD~\cite{vqarad}      &  2   \\

\bottomrule
\end{tabular}%
\caption{Statistics of VerMulti.}
\label{tab:mathv60k_statistics}
\end{table}

\noindent\textbf{VerMulti-Geo15K:} We use 15k geometry problems from MathV360K~\cite{mathv} to train MGT-Geo, which consists with GEOS~\cite{geos}, Geometry3K~\cite{geometry3k}, GeoQA+~\cite{geoqa+}, UniGeo~\cite{unigeo}, and TQA~\cite{tqa}. The distribution of data is shown in \cref{tab:mathv_geo_statistics}.

\noindent\textbf{Sokoban Training Datasets:} For the Sokoban domain, we generated a diverse set of 11,500 environments by using gym-sokoban~\cite{sokoban} with varying difficulty levels to train MGT-Sokoban. Specifically, we created 5,000 environments using the sokoban-small-v0 difficulty setting and another 5,000 environments with sokoban-small-v1 difficulty. To ensure the model encounters a wide range of scenarios, we further supplemented the dataset with 500 environments each from sokoban-v0, sokoban-v1, and sokoban-large-v0 difficulty settings. This distribution allows the model to learn planning strategies across different board sizes and complexity levels, from simpler small-scale puzzles to more challenging large configurations.

\begin{table}[!h]
\centering
\setlength\tabcolsep{15pt}
\begin{tabular}{c|c}
\toprule
Split      & Count \\ \midrule
GEOS       & 271   \\
GeoQA+     & 8155  \\
Geometry3K & 2776  \\
TQA        & 25    \\
UniGeo     & 5583  \\ \bottomrule
\end{tabular}%
\caption{Statistics of VerMulti-Geo15K.}
\label{tab:mathv_geo_statistics}
\end{table}

\subsection{Training Hyper-Parameters}
Our training infrastructure is built upon OpenRLHF~\cite{openrlhf}, and we use the same set of hyper-parameters for RL, as shown in \cref{tab:hyper-parameters}.

We train SFT-Text 2 epochs with batch-size=256, learning-rate=5e-6, warmup-ratio=0.03.

\section{Detailed Performance on MM-Star}
To validate the improvement of our model on general visual tasks, we also provide the MM-Star different split results in \cref{tab:mm_star_split_results}. The results show that MGT can improve the perception ability of FRE-Text.

\begin{table}[!h]
\centering
\setlength{\tabcolsep}{6.5pt}
\begin{tabular}{l|ccc}
    \toprule
    \textbf{Model} & \textbf{Coarse} & \textbf{Fine-grained} & \textbf{AVG} \\
    \midrule
    QwenVL-2.5 CoT  & 68.56 & 47.52 & 58.04 \\
    FRE-Multi  & 69.32 & 52.05 & 60.68 \\
    FRE-Text  & 65.20 & 46.30 & 55.75 \\
    MGT-PerceReason  & 67.99 & 48.19 & 58.09 \\
    \bottomrule
\end{tabular}
\caption{Performance comparison (\%) on different perception categories in MM-Star benchmark. Coarse Perception includes tasks requiring basic object recognition, while Fine-grained Perception involves more detailed visual discrimination tasks.}
\label{tab:mm_star_split_results}
\end{table}

\section{Evaluation Details}
\subsection{Text-Only Benchmarks}
We use LightEval~\cite{lighteval} to evaluate our models on text-only mathematical reasoning benchmarks. LightEval is a lightweight evaluation framework that provides standardized evaluation protocols for language models. For our evaluation, we focus on two challenging mathematical reasoning benchmarks: MATH-500~\cite{stepbystep} and GPQA-Diamond~\cite{gpqa}.

\paragraph{Evaluation Protocol.} 
We implement custom evaluation tasks within the LightEval~\cite{lighteval} framework to ensure consistent and reproducible evaluation. For each benchmark, we define specific prompt templates that encourage step-by-step reasoning and clear answer formatting, as shown in \cref{app:prompts}. The evaluation is conducted with a maximum generation length of 32,768 tokens to accommodate extensive reasoning chains.

\paragraph{Metrics.} 
For MATH-500, we employ a specialized metric that extracts answers from LaTeX expressions, with priority given to boxed answers. This metric uses multilingual extractive matching with a precision of 5 decimal places. For GPQA-Diamond, we use a letter-based extraction metric that identifies the selected multiple-choice option (A, B, C, or D) from the model response.

Both metrics are implemented using LightEval's multilingual extractive match framework, which provides robust answer extraction capabilities even when models deviate slightly from the requested output format. This approach ensures fair comparison across different model architectures and response styles.

\begin{table*}[ht]
\centering
\begin{tabular}{@{}rccccc@{}}
\toprule
                        & FRE-Text & FRE-Multi & MGT-Geo & MGT-PerceReason & MGT-Sokoban\\ \midrule
train\_batch\_size      & 128       & 256 & 256 & 256   & 128           \\
rollout\_batch\_size    & 128       & 256 & 256 & 256      & 128        \\
temperature             & 1         & 1 & 1 & 1        & 1      \\
n\_samples\_per\_prompt & 16        & 16 & 16  & 16     & 16       \\
max\_epochs             & 1         & 1 & 1   & 1      &1     \\
num\_episodes           & 1         & 2 & 2   & 2    & 4       \\
generate\_max\_len      & 8192      & 8192 & 8192 & 8192    & 8192         \\
init\_kl\_coef          & 1e-3      & 1e-3 & 1e-3   & 1e-3       & 0.0     \\
lambd                   & 1         & 1 & 1   & /     & 1       \\
gamma                   & 1         & 1 & 1   & /      & 1       \\
actor\_learning\_rate   & 1e-6      & 1e-6 & 4e-7 & 4e-7       & 1e-6       \\
critic\_learning\_rate  & 9e-6      & 9e-6 & 9e-6   & /      & 5e-6      \\ 
warm-up ratio           & 0.03      & 0.03 & 0.03 & 0.03    & 0.03       \\ \bottomrule
\end{tabular}
\caption{Training Hyper-parameters.}
\label{tab:hyper-parameters}
\end{table*}

\subsection{Multimodal Benchmark}
We use LMMs-Eval~\cite{lmmseval} for multimodal benchmark evaluation, which is a comprehensive and lightweight evaluation toolkit for LMMs. We evaluate our models in representative challenging multimodal reasoning benchmarks, including MathVision~\cite{mathvision}, MathVerse~\cite{mathverse}, MathVista~\cite{mathvista}, OlympiadBench~\cite{olympiadbench} and MM-Star~\cite{mmstar}.

\paragraph{Evaluation Protocol.}
We integrate vLLM~\cite{vllm} into LMMs-Eval and deploy our models on vLLM for efficient evaluation. To accurately extract the answer of the model for each test case, we deploy Qwen2.5-14B-Instruct~\cite{qwen2.5} as a judge, using the prompt shown in \cref{app:prompts}.

The system prompts of our models used in training and evaluation are the same, which are also listed in \cref{app:prompts}. Furthermore, we find that using the same system prompt for the baseline model degrades its performance. Thus, for fairness, we do not use our system prompt for the baseline model.

\subsection{Agent-Related Benchmark}
We follow the MageBench~\cite{magebench} to evaluate our model in Sokoban and football tasks. 

\paragraph{Agent Formulation.} 
We use two agent designs from MageBench~\cite{magebench} for our evaluations:

\noindent\textbf{Global Planner Agent:} This agent observes the initial environment once and plans all subsequent actions without further observations. Its decision process is:
\begin{equation}
    \pi_\theta (p_{sys}, p_{task}, p_{cot}, p_{io}) \rightarrow a_1, a_2, ..., a_T
\end{equation}
    
\noindent\textbf{Online Planner Agent:} This agent analyzes each step and acts based on previous actions (AM) and observations (OM). We set AM = 5 and OM = 1. Its decision process is:
\begin{equation}
    \pi_\theta (p_{sys}, a_{t-AM:t}, o_{t-OM:t}, p_{cot}, p_{io}) \rightarrow a_{t+1}
\end{equation}

\paragraph{Task Descriptions.}
\begin{itemize}
    \item \textbf{Sokoban-Global:} The model sees the initial puzzle once and plans all moves in advance. This tests the model's ability to mentally simulate the environment.
    
    \item \textbf{Football-Online:} The model controls a player in a dynamic environment, making decisions based on changing game states. This tests real-time decision-making in complex scenarios.
\end{itemize}

\clearpage
\onecolumn


\section{Training Curves}
\begin{figure}[!h]
    \centering
   \subfloat[]{
       \includegraphics[width=0.35\textwidth]{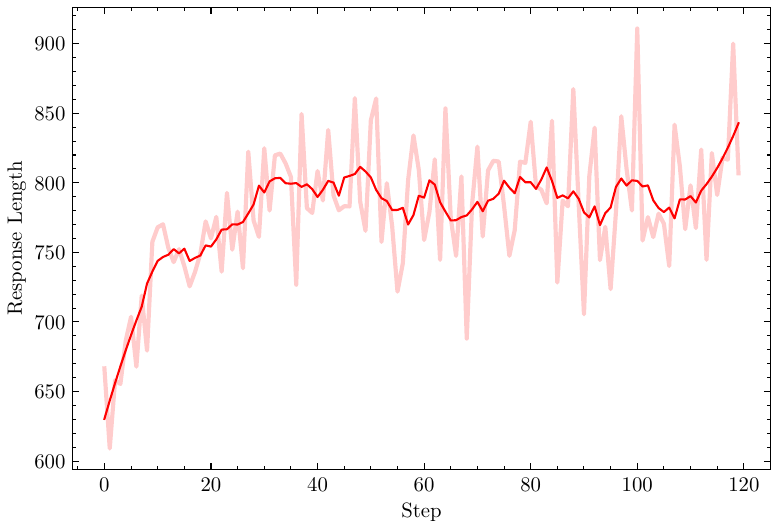}
   }
   \subfloat[]{
       \includegraphics[width=0.35\textwidth]{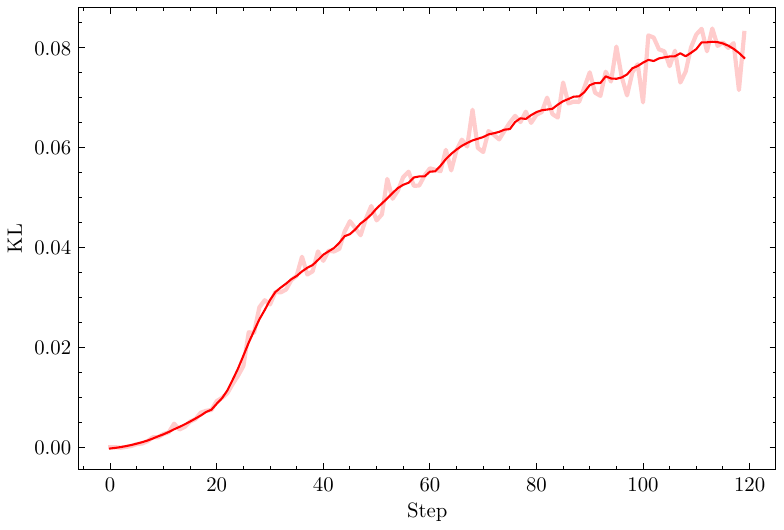}
   }
   \\
   \subfloat[]{
       \includegraphics[width=0.35\textwidth]{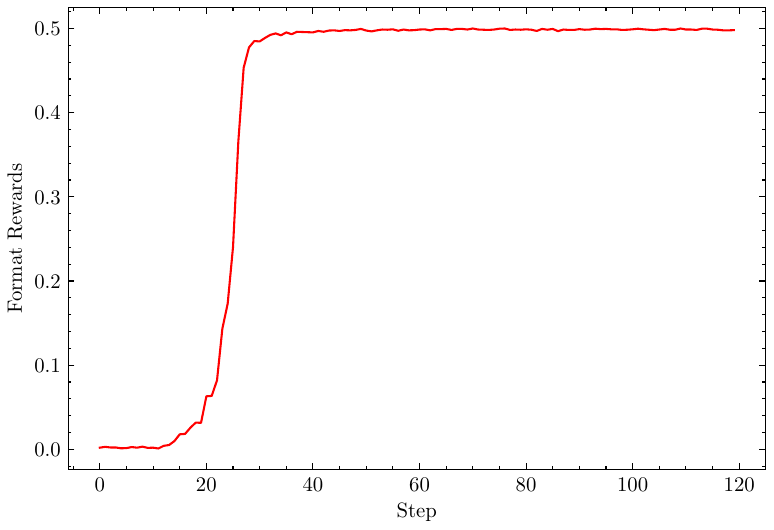}
   }
   \subfloat[]{
       \includegraphics[width=0.35\textwidth]{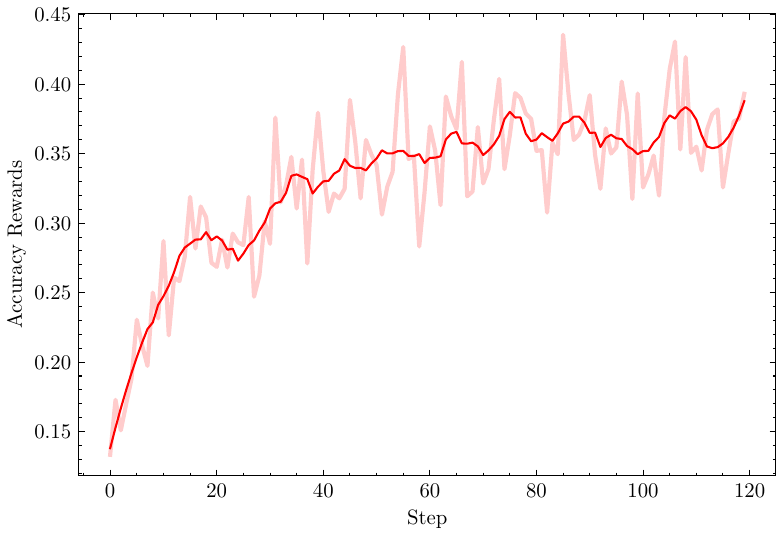}
   }
   \vspace{-5pt}
   \caption{Training Curve of FRE-Text.}
   \label{fig:3b_train_curve}
\end{figure}
\begin{figure}[!h]
    \centering
   \subfloat[]{
       \includegraphics[width=0.35\textwidth]{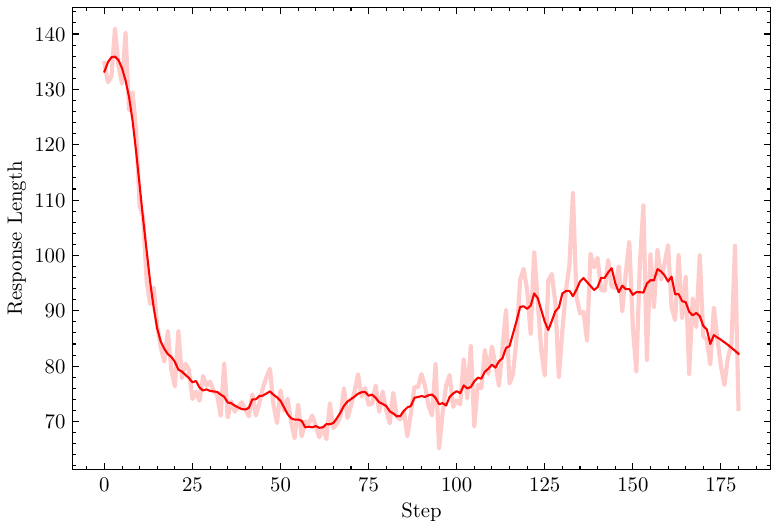}
   }
   \subfloat[]{
       \includegraphics[width=0.35\textwidth]{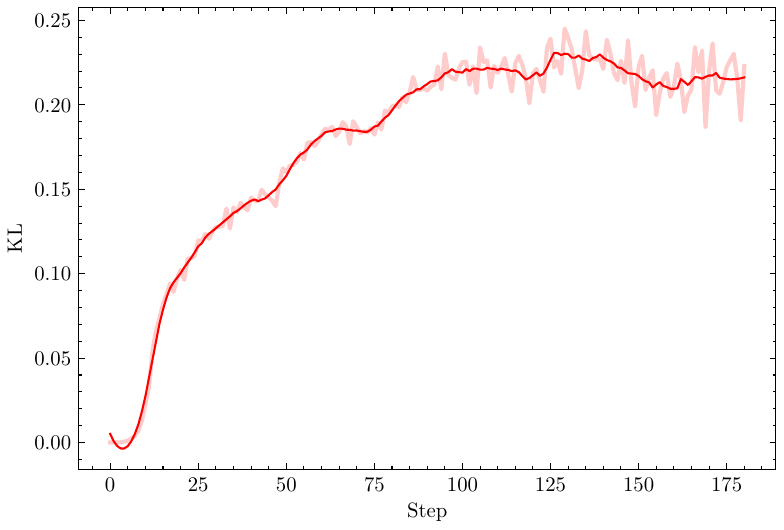}
   }
   \\
   \subfloat[]{
       \includegraphics[width=0.35\textwidth]{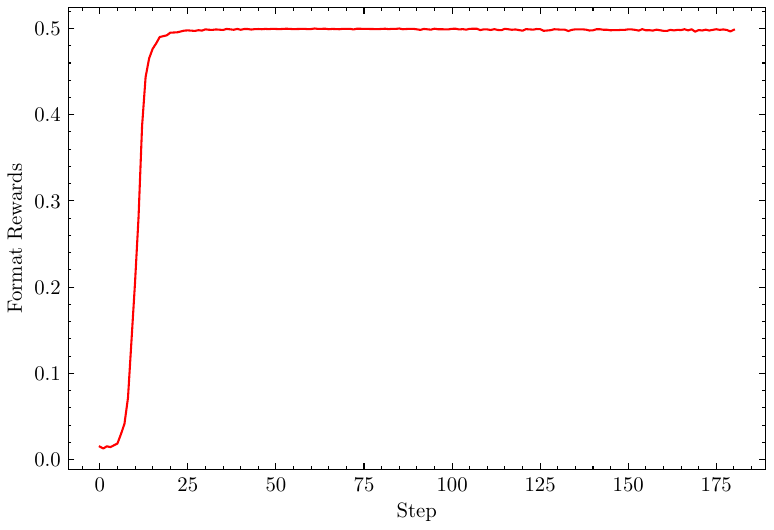}
   }
   \subfloat[]{
       \includegraphics[width=0.35\textwidth]{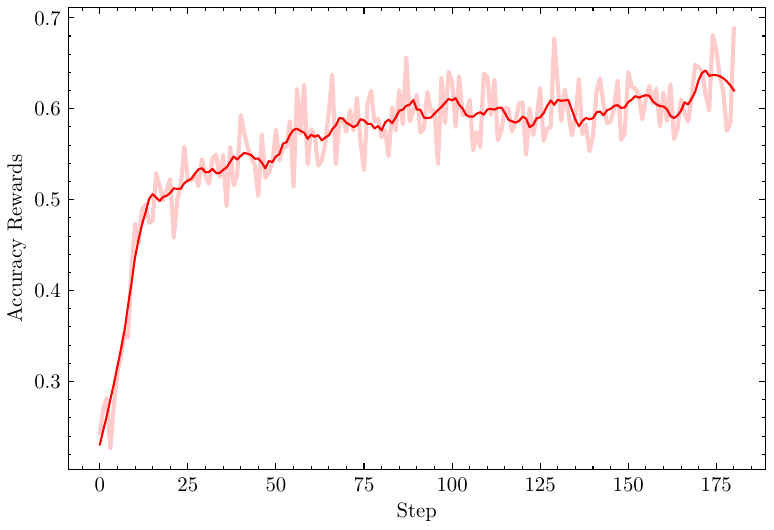}
   }
   \vspace{-5pt}
   \caption{Training Curve of FRE-Multi.}
   \label{fig:3b_mathv50_train_curve}
\end{figure}

\begin{figure}[!h]
    \centering
   \subfloat[]{
       \includegraphics[width=0.35\textwidth]{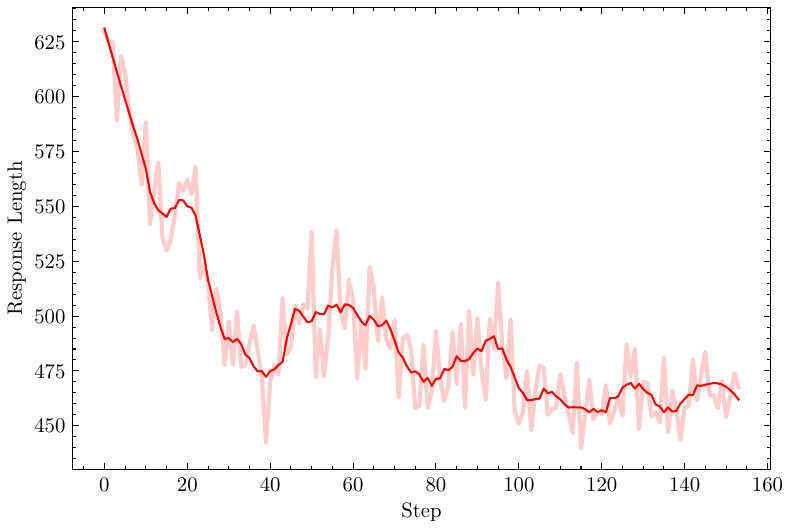}
   }
   \subfloat[]{
       \includegraphics[width=0.35\textwidth]{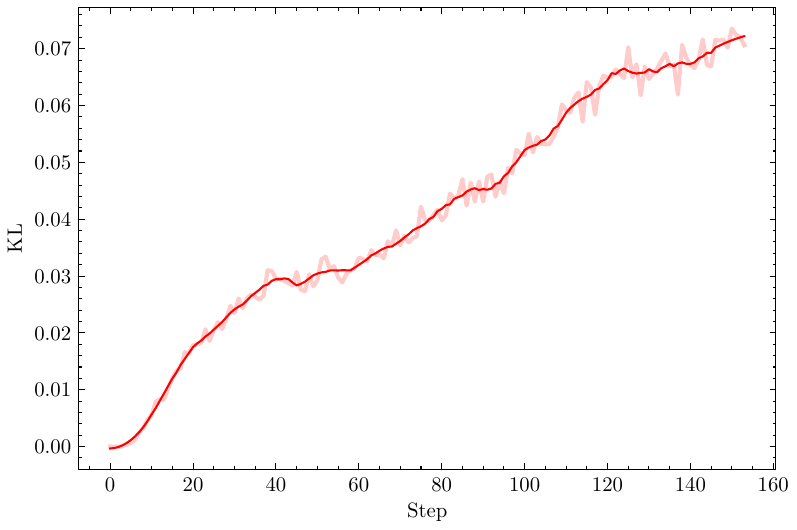}
   }
   \\
   \subfloat[]{
       \includegraphics[width=0.35\textwidth]{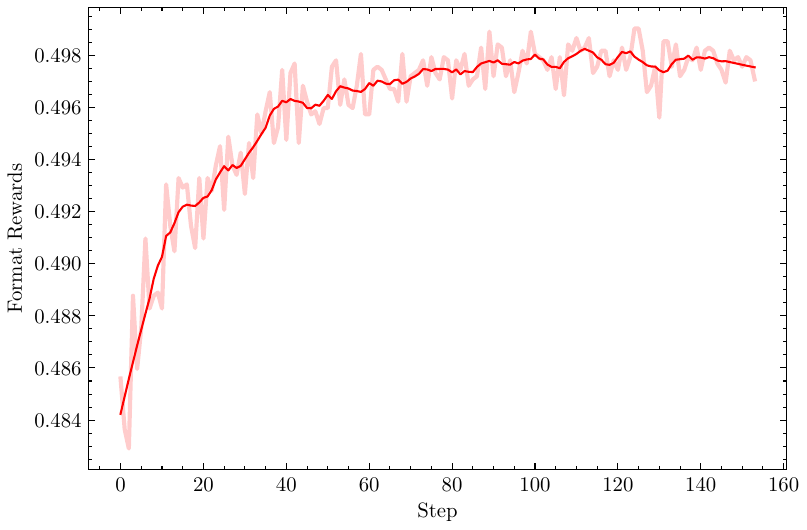}
   }
   \subfloat[]{
       \includegraphics[width=0.35\textwidth]{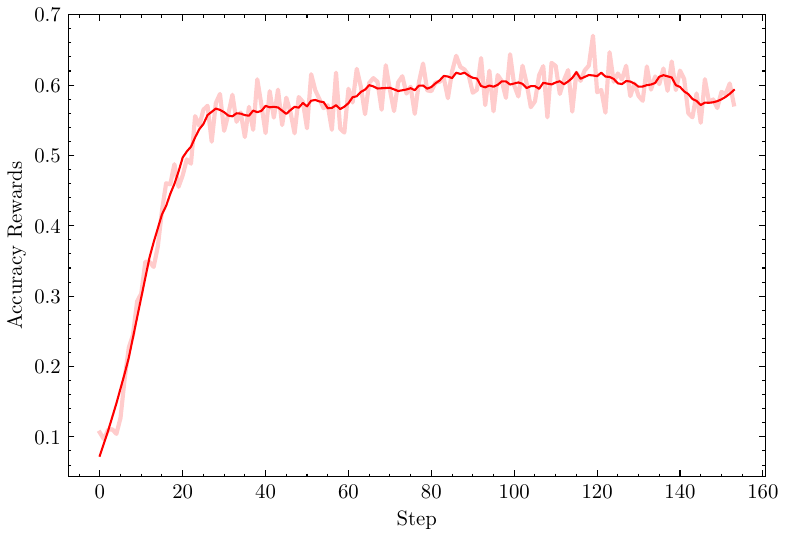}
   }
   \vspace{-5pt}
   \caption{Training Curve of MGT-Geo.}
   \label{fig:3b_geo_train_curve}
\end{figure}

\begin{figure}[!h]
    \centering
   \subfloat[]{
       \includegraphics[width=0.35\textwidth]{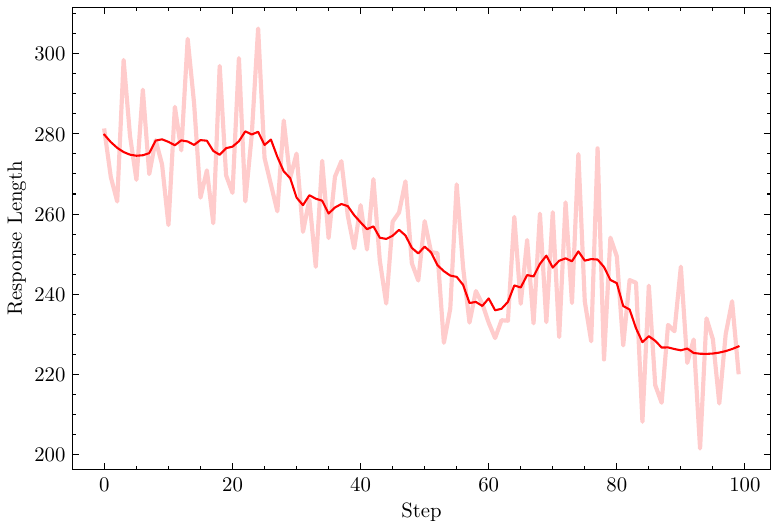}
   }
   \subfloat[]{
       \includegraphics[width=0.35\textwidth]{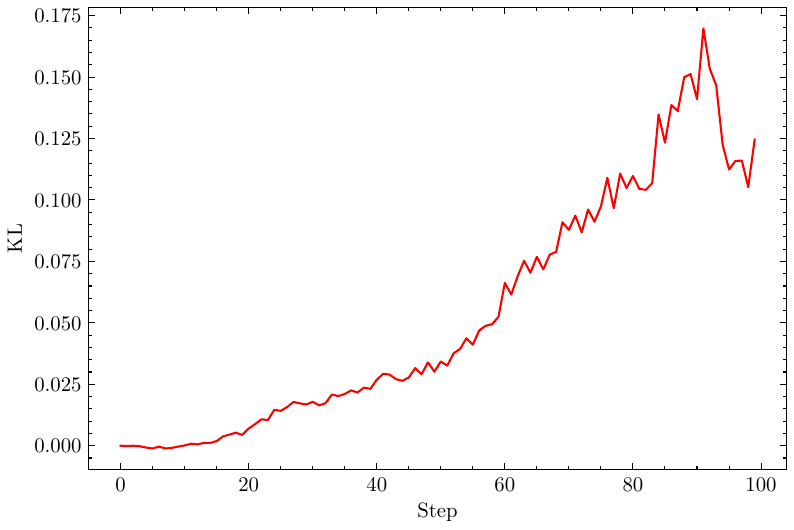}
   }
   \\
   \subfloat[]{
       \includegraphics[width=0.35\textwidth]{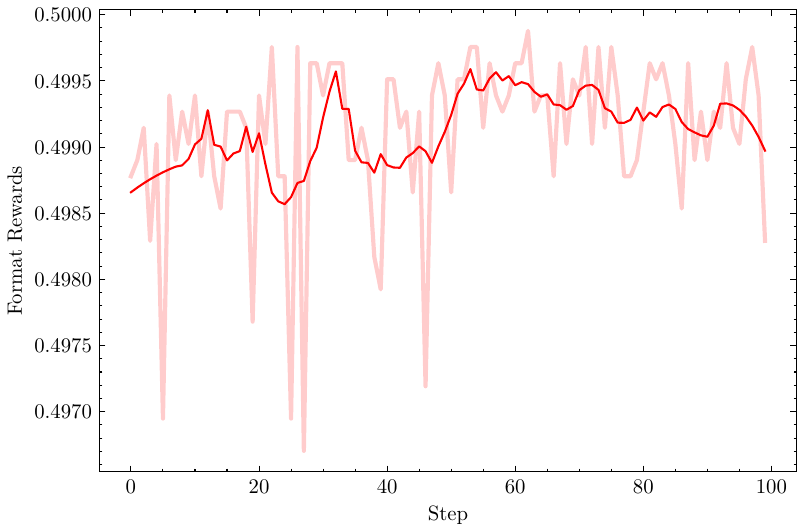}
   }
   \subfloat[]{
       \includegraphics[width=0.35\textwidth]{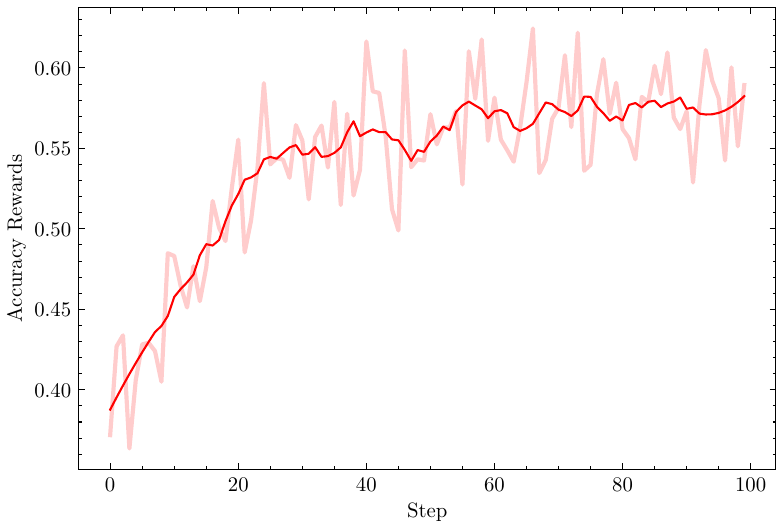}
   }
   \vspace{-5pt}
   \caption{Training Curve of MGT-PerceReason.}
   \label{fig:3b_mathv50_after_text_train_curve}
\end{figure}

\clearpage

\section{Prompts for Training and Evaluation}
\label{app:prompts}
\begin{tcolorbox}[breakable,title={System Prompts for Training and Evaluation}]
\begin{lstlisting}
You are a helpful assistant good at solving math problems with step-by-step reasoning. You should first thinks about the reasoning process in the mind and then provides the user with the answer. Your answer must be in latex format and wrapped in $...$.The reasoning process and answer are enclosed within <think> </think> and <answer> </answer> tags, respectively, i.e., <think> Since $1+1=2$, so the answer is $2$. </think><answer> $2$ </answer>, which means your output should start with <think> and end with </answer>.
\end{lstlisting}
\end{tcolorbox}
\begin{tcolorbox}[breakable,title={Prompt Templates for Text-Only Benchmarks}]
\paragraph{MATH-500 Prompt Template:}
\begin{lstlisting}
Solve the following math problem efficiently and clearly. The last line of your response should be of the following format: 'Therefore, the final answer is: $\boxed{ANSWER}$. I hope it is correct' (without quotes) where ANSWER is just the final number or expression that solves the problem. Think step by step before answering.

{Question}
\end{lstlisting}

\paragraph{GPQA-Diamond Prompt Template:}
\begin{lstlisting}
Answer the following multiple choice question. The last line of your response should be of the following format: 'Answer: $LETTER' (without quotes) where LETTER is one of ABCD. Think step by step before answering.

{Question}

A) {A}
B) {B}
C) {C}
D) {D}
\end{lstlisting}
\end{tcolorbox}
\begin{tcolorbox}[breakable,title={Prompt of Answer Extraction for Qwen2.5-14B-Instruct}]
\begin{lstlisting}
Please read the following example. Then extract the answer from the model response and type it at the end of the prompt.

Hint: Please answer the question requiring an integer answer and provide the final value, e.g., 1, 2, 3, at the end.
Question: Which number is missing?

Model response: The number missing in the sequence is 14.

Extracted answer: 14

Hint: Please answer the question requiring a floating-point number with one decimal place and provide the final value, e.g., 1.2, 1.3, 1.4, at the end.
Question: What is the fraction of females facing the camera?

Model response: The fraction of females facing the camera is 0.6, which means that six out of ten females in the group are facing the camera.

Extracted answer: 0.6

Hint: Please answer the question requiring a floating-point number with two decimal places and provide the final value, e.g., 1.23, 1.34, 1.45, at the end.
Question: How much money does Luca need to buy a sour apple candy and a butterscotch candy? (Unit: $)

Model response: Luca needs $1.45 to buy a sour apple candy and a butterscotch candy.

Extracted answer: 1.45

Hint: Please answer the question requiring a Python list as an answer and provide the final list, e.g., [1, 2, 3], [1.2, 1.3, 1.4], at the end.
Question: Between which two years does the line  graph saw its maximum peak?

Model response: The line graph saw its maximum peak between 2007 and 2008.

Extracted answer: [2007, 2008]

Hint: Please answer the question and provide the correct option letter, e.g., A, B, C, D, at the end.
Question: What fraction of the shape is blue?\nChoices:\n(A) 3/11\n(B) 8/11\n(C) 6/11\n(D) 3/5

Model response: The correct answer is (B) 8/11.

Extracted answer: B

Hint: Please answer the question and provide the correct option letter, e.g., A, B, C, D, at the end.
Question: {question}

Model response: {response}

Extracted answer:
\end{lstlisting}
\end{tcolorbox}

\section{Case Study}
\label{app:case_study}
\begin{tcolorbox}[breakable,title={Case Study of Qwen2.5-VL CoT and FRE-Text}]
Question: Find the sum of all real solutions to the equation $\frac{x-2}{x^2+4x+1} = \frac{x-5}{x^2-10x}$.

------------------------------------------------------------------------------------------------------------------

Qwen2.5-VL CoT:

\begin{lstlisting}
Rearrange equation: divide by x^2 and let's use partial fractions to simplify the equation.
To simplify, write each fraction as a difference: $1 + \frac{x-4}{(x+2)(x+2)(x-5)} = 1 + \frac{(x-2) - (x-5)}{x^2-10x}$.
3. 2x^2 - 4x - 16 = 0.
4. Add 16 to both sides: 2x^2 - 4x = 0
5. Divide by 2: x^2 - 2x = 0.
6. Take out x: x(x-2) = 0. Elie: x = 0 or x = 2.
Therefore, the sum of all real solutions to the equation is x = 0 + 2 = $\boxed{2}$.
\end{lstlisting}

------------------------------------------------------------------------------------------------------------------

FRE-Text:

\begin{lstlisting}
<think>
To solve the equation \(\frac{x-2}{x^2+4x+1} = \frac{x-5}{x^2-10x}\), we first eliminate the denominators by cross-multiplying:

\[
(x-2)(x^2-10x) = (x-5)(x^2+4x+1)
\]

Expanding both sides, we get:

\[
x^3 - 10x^2 - 2x^2 + 20x = x^3 + 4x^2 + x - 5x^2 - 20x - 5
\]

Simplifying both sides, we have:

\[
x^3 - 12x^2 + 20x = x^3 - x^2 - 19x - 5
\]

Subtracting \(x^3\) from both sides, we get:

\[
-12x^2 + 20x = -x^2 - 19x - 5
\]

Rearranging all terms to one side, we have:

\[
-12x^2 + x^2 + 20x + 19x + 5 = 0
\]

Simplifying further, we get:

\[
-11x^2 + 39x + 5 = 0
\]

This is a quadratic equation in the form \(ax^2 + bx + c = 0\), where \(a = -11\), \(b = 39\), and \(c = 5\). The sum of the solutions of a quadratic equation \(ax^2 + bx + c = 0\) is given by \(-\frac{b}{a}\). Therefore, the sum of the solutions is:

\[
-\frac{39}{-11} = \frac{39}{11}
\]

We need to check if the solutions satisfy the original equation and the denominators. The solutions to the quadratic equation are:

\[
x = \frac{-39 \pm \sqrt{39^2 - 4(-11)(5)}}{2(-11)} = \frac{-39 \pm \sqrt{1521 + 220}}{-22} = \frac{-39 \pm \sqrt{1741}}{-22}
\]

The solutions are:

\[
x = \frac{39 \mp \sqrt{1741}}{22}
\]

We need to ensure that these solutions do not make the denominators zero. The denominators are \(x^2 + 4x + 1\) and \(x^2 - 10x\). Substituting \(x = \frac{39 \mp \sqrt{1741}}{22}\) into these denominators, we check:

For \(x^2 + 4x + 1\):

\[
\left(\frac{39 \mp \sqrt{1741}}{22}\right)^2 + 4\left(\frac{39 \mp \sqrt{1741}}{22}\right) + 1 \neq 0
\]

For \(x^2 - 10x\):

\[
\left(\frac{39 \mp \sqrt{1741}}{22}\right)^2 - 10\left(\frac{39 \mp \sqrt{1741}}{22}\right) \neq 0
\]

Since the solutions do not make the denominators zero, they are valid. Therefore, the sum of all real solutions is:

\[
\boxed{\frac{39}{11}}
\]
</think><answer> $\frac{39}{11}$ </answer>
\end{lstlisting}

\end{tcolorbox}

\begin{tcolorbox}[breakable,title={Case Study of Qwen2.5-VL CoT and FRE-Text}]
Question: How many positive integers $b$ have the property that $\log_{b} 729$ is a positive integer?

------------------------------------------------------------------------------------------------------------------

Qwen2.5-VL CoT:

\begin{lstlisting}
We need to find positive integers $b$ such that $\log_{b} 729$ is a positive integer. This means that $b^{\log_{b} 729} = 729$.

Let's find the prime factorization of 729. We have $729 = 3^6$. 

We know that if $\log_{b} 3^6$ is a positive integer, then $b$ must be a power of 3. Let's write $b = 3^x$. Then we have $\log_{3^x} 3^6 = 2$ since $b^q = 3^6$ implies that $3^{x \cdot \log_{3} 3^6} = 3^6$ and $\log_{3} 3^6 = 6$. Thus, we have $x \cdot \log_{3} 3^6 = 6$. This simplifies to $x \cdot 6 = 6$, so $x = 1$. 

This means $b = 3^1 = 3$ is a solution. We need to check other powers of 3, but since $\log_{3} 3^6 = 6$, any other power of 3 greater than 3 would result in a non-positive integer for $\log_{b} 729$. Therefore, there is only one positive integer $b$ for which $\log_{b} 729$ is a positive integer.

The answer is: 1
\end{lstlisting}

------------------------------------------------------------------------------------------------------------------

FRE-Text:

\begin{lstlisting}
<think>
To determine how many positive integers \( b \) have the property that \(\log_{b} 729\) is a positive integer, we start by letting \(\log_{b} 729 = k\), where \( k \) is a positive integer. This equation can be rewritten in its exponential form as:

\[ b^k = 729 \]

We need to find all possible values of \( b \) such that \( 729 \) is a perfect power of \( b \). First, we factorize \( 729 \):

\[ 729 = 3^6 \]

For \( b^k = 3^6 \) to hold, \( b \) must be a power of \( 3 \). Let \( b = 3^m \), where \( m \) is a positive integer. Then we have:

\[ (3^m)^k = 3^{6k} \]

This implies that \( m \cdot k = 6 \). We need to find all pairs \((m, k)\) such that \( m \cdot k = 6 \). The possible pairs \((m, k)\) are:

1. \( (1, 6) \)
2. \( (2, 3) \)
3. \( (3, 2) \)
4. \( (6, 1) \)

For each pair \((m, k)\), we can find \( b \):

1. If \( m = 1 \) and \( k = 6 \), then \( b = 3^1 = 3 \).
2. If \( m = 2 \) and \( k = 3 \), then \( b = 3^2 = 9 \).
3. If \( m = 3 \) and \( k = 2 \), then \( b = 3^3 = 27 \).
4. If \( m = 6 \) and \( k = 1 \), then \( b = 3^6 = 729 \).

Thus, the possible values of \( b \) are \( 3, 9, 27, \) and \( 729 \). There are 4 such values.

Therefore, the number of positive integers \( b \) that have the property that \(\log_{b} 729\) is a positive integer is:\[\boxed{4}\]</think><answer> $4$ </answer>
\end{lstlisting}

\end{tcolorbox}


\begin{tcolorbox}[breakable,title={Case Study of Qwen2.5-VL and FRE-Multi}]
Question: What is the median number of points scored by the team per game?

\includegraphics[width=0.2\columnwidth]{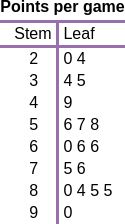}


------------------------------------------------------------------------------------------------------------------

Qwen2.5-VL:

\begin{lstlisting}
<think>
Step 1: Identify all the data points in the stem and leaf plot.
From the stem and leaf plot, the points per game are as follows:
- 21
- 24
- 34, 35
- 49
- 56, 57, 58
- 66, 66
- 75, 76
- 84, 85, 85, 85
- 90

Step 2: Arrange the data points in ascending, descending or random order.
The points in ascending order are:
21, 24, 34, 35, 49, 56, 57, 58, 66, 66, 75, 76, 84, 85, 85, 85, 90

Step 3: Find the median.
Since there are 17 data points (an odd number), the median is the middle number when ordered.
In this case, the 9th number in the ordered list is the median.
The 9th number is 56.

Therefore, the median number of points scored per game is 56.

</think><answer> $\\boxed{56}$ </answer>
\end{lstlisting}

------------------------------------------------------------------------------------------------------------------

FRE-Multi:

\begin{lstlisting}
<think>
To find the median, I need to follow these steps:
1. List all the values in ascending order.\n2. Identify the middle value(s).

Let's list the values:\n20, 24, 20, 30, 34, 35, 40, 49, 50, 56, 57, 58, 60, 60, 66, 66, 70, 75, 76, 80, 85, 85, 85, 86, 90

There are 25 values, so the median will be the 13th value in this ordered list.
The 13th value is 60.

Therefore, the median number of points scored per game is 60.
</think><answer> $\\boxed{60}$ </answer>
\end{lstlisting}

\end{tcolorbox}

\begin{tcolorbox}[breakable,title={Case Study of Qwen2.5-VL v.s. FRE-Multi}]
Question: What is the value of `legs' for `grade'? \\
\includegraphics[width=0.3\columnwidth]{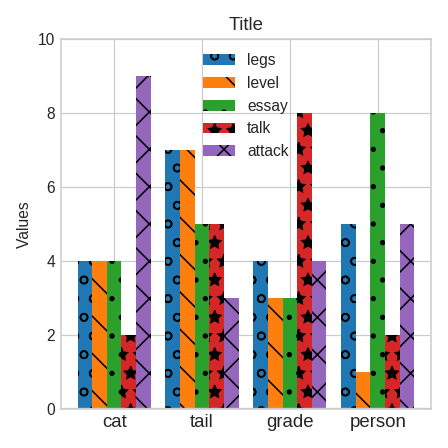}

-----------------------------------------------------------------------------------------------------------------

Qwen2.5-VL:

\begin{lstlisting}
<think> To find the value of 'legs' for 'grade', we need to look at the legend of the graph, which clearly indicates that 'legs' are represented by the blue circles. After identifying the data for \"grade\" in the graph, we see that the corresponding value for 'legs' is 4. Therefore, the value of 'legs' for 'grade' is 4.\n</think><answer> $\\boxed{4}$ </answer>
\end{lstlisting}

------------------------------------------------------------------------------------------------------------------

FRE-Multi:

\begin{lstlisting}
<think> The 'legs' value for 'grade' is shown on the bar chart as 4. </think><answer> $\\boxed{4}$ </answer>
\end{lstlisting}

\end{tcolorbox}

\clearpage

\begin{tcolorbox}[breakable,title={Case Study of Qwen2.5-VL and FRE-Multi}]
Question: How many vehicles in the image have wheels?

\includegraphics[width=0.3\columnwidth]{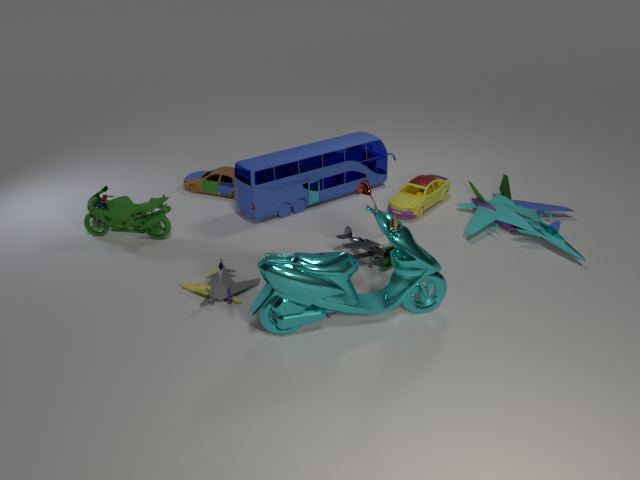}

------------------------------------------------------------------------------------------------------------------

Qwen2.5-VL:

\begin{lstlisting}
<think> 
First, I'll identify each vehicle in the image. There is a double bus (blue), a car (tan), a car (yellow), a motorcycle (green), a scooter (cyan), and an airplane (gray). Some of these are likely to have wheels. 

Now, examining the vehicles one by one:
- The blue double bus typically has wheels.
- The tan car and the yellow car have wheels.
- The green motorcycle has two wheels.
- The scooter likely has two wheels as well.
- The gray airplane typically has small wheels beneath.

When considering this categorically, all of these vehicles seem to have wheels. However, classifying vehicles as always having wheels or as having potential wheels introduces subjectivity. Typically most vehicles have wheels unless otherwise specified, so I'll make conservative assumptions regarding certainty.

I'll perform a count suggesting the vehicles have wheels:
1. The double bus has wheels.
2. The cars each usually have wheels.
3. The motorcycle presumably has wheels.
4. The scooter likely has typical two wheels.
5. The airplane typically has wheels.\n\nThus, I will count the potentially wheel-related vehicles (assumed to have wheels before explicit confirmation):
1. (Wheel-like others might exist)\n\nFinal number of vehicles likely to have wheels:
\\boxed{4}
</think><answer> $\\boxed{4}$ </answer>
\end{lstlisting}

------------------------------------------------------------------------------------------------------------------

FRE-Multi:

\begin{lstlisting}
<think>
We need to count the number of vehicles with wheels in the image. Let's identify them:

1. The blue bus has wheels.
2. The green motorcycle has wheels.
3. The brown sedan has wheels.
4. The yellow sedan has wheels.
5. The cyan scooter has wheels.

There are no other vehicles with wheels in the image.

Therefore, the total number of vehicles with wheels is 5.
</think><answer> $\\boxed{5}$ </answer>
\end{lstlisting}

\end{tcolorbox}

\clearpage
\begin{tcolorbox}[breakable,title={Case Study of Qwen2.5-VL and FRE-Multi}]
Question: What is the purpose of the left lane in the picture? 

Choices:

(A) To show the results of immunofluorescent labeling 

(B) To indicate the upper layer of synovial membranes 

(C) To show the magnification of the image

(D) To display the results of immunohistochemistry

\includegraphics[width=0.3\columnwidth]{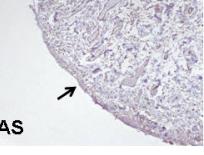}

------------------------------------------------------------------------------------------------------------------

Qwen2.5-VL:

\begin{lstlisting}
 <think> The image appears to be a microscope slide of tissue sections. The left lane consistently showing up in the micrograph suggests that it is used to display the results of immunohistochemistry, where antibodies are applied to stained tissue sections to detect specific molecules. The other options do not align with the visual presentation and technical names associated with immunohistochemistry. </think><answer> $\\boxed{D}$ </answer>
\end{lstlisting}

------------------------------------------------------------------------------------------------------------------

FRE-Multi:

\begin{lstlisting}
<think> The presence of the arrow and the context suggest that the left lane is used to display the results of immunohistochemistry. </think><answer> \\boxed{D} </answer>

\end{lstlisting}

\end{tcolorbox}

\end{document}